\documentclass{article}
\newcommand{\ignore}[1]{}
\usepackage{tikz}
\newcommand*\circled[1]{\tikz[baseline=(char.base)]{
            \node[shape=circle,draw,inner sep=1pt] (char) {#1};}}

\usepackage{microtype}
\usepackage{graphicx}
\usepackage{subfigure}
\usepackage{booktabs} % for professional tables
\usepackage{enumitem}

\usepackage{hyperref}

\usepackage[accepted]{icml2024}

\usepackage{amsmath}
\usepackage{amssymb}
\usepackage{mathtools}
\usepackage{amsthm}

\DeclareMathOperator*{\argmin}{arg\,min}
\usepackage{mathtools}

\usepackage[capitalize,noabbrev]{cleveref}

\theoremstyle{plain}

\theoremstyle{definition}

\theoremstyle{remark}

\usepackage[textsize=tiny]{todonotes}
 
\icmltitlerunning{Progressive Inference: Explaining Decoder-Only Sequence Classification Models Using Intermediate Predictions}

\begin{document}

\twocolumn[
\icmltitle{Progressive Inference: Explaining Decoder-Only Sequence Classification Models Using Intermediate Predictions}




\begin{icmlauthorlist}
\icmlauthor{Sanjay Kariyappa}{jpm}
\icmlauthor{Freddy Lécué}{jpm}
\icmlauthor{Saumitra Mishra}{jpm}
\icmlauthor{Christopher Pond}{jpm}
\icmlauthor{Daniele Magazzeni}{jpm}
\icmlauthor{Manuela Veloso}{jpm}
\end{icmlauthorlist}

\icmlaffiliation{jpm}{JPMorganChase AI Research}

\icmlcorrespondingauthor{Sanjay Kariyappa}{sanjay.kariyappa@jpmchase.com}

\icmlkeywords{Machine Learning, ICML}

\vskip 0.3in
]



\printAffiliationsAndNotice{}  % leave blank if no need to mention equal contribution

\begin{abstract}
This paper proposes \emph{Progressive Inference}--a framework to compute input attributions to explain the predictions of decoder-only sequence classification models. Our work is based on the insight that the classification head of a decoder-only Transformer model can be used to make \emph{intermediate predictions} by evaluating them at different points in the input sequence. Due to the causal attention mechanism, these intermediate predictions only depend on the tokens seen before the inference point, allowing us to obtain the model's prediction on a masked input sub-sequence, with negligible computational overheads. We develop two methods to provide sub-sequence level attributions using this insight. First, we propose \emph{Single Pass-Progressive Inference (SP-PI)}, which computes attributions by taking the difference between consecutive intermediate predictions. Second, we exploit a connection with Kernel SHAP to develop \emph{Multi Pass-Progressive Inference (MP-PI)}. MP-PI uses intermediate predictions from multiple masked versions of the input to compute higher quality attributions. Our studies on a diverse set of models trained on text classification tasks show that SP-PI and MP-PI provide significantly better attributions compared to prior work. 
\end{abstract}
\begin{figure}[t]
  \centering
  \includegraphics[width=0.8\columnwidth]{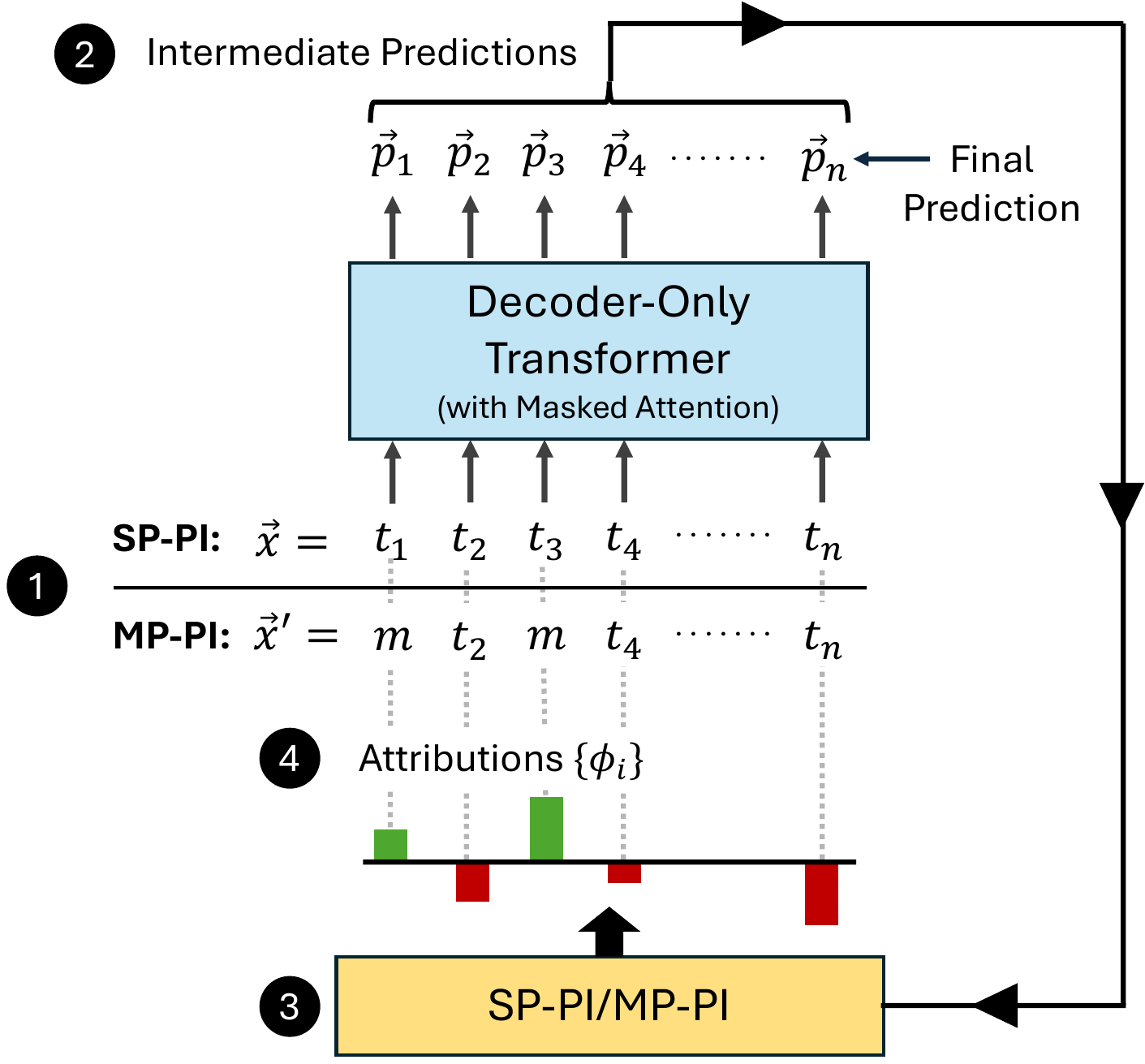}
  \vspace{-0.1in}
  \caption{\circled{1} \emph{Input tokens} are fed to the decoder-only models to produce \circled{2} \emph{intermediate predictions}. \circled{3} \emph{Progressive inference (PI)} uses these predictions to produce \circled{4} \emph{attributions} over input tokens/words/sentences. While \emph{Single-Pass PI} uses the intermediate predictions produced by the original input tokens, \emph{multi-pass PI} collects multiple sets of intermediate predictions with different masked versions of the input to compute the attribution.}
  \label{fig:overview}
\end{figure}

\section{Introduction}
Large language Models (LLMs) based on the decoder-only Transformer architecture~\cite{vaswani2017attention} (e.g. GPT~\cite{radford2018improving}) have gained widespread adoption over the past few years with a burgeoning open-source community creating increasingly performant models. Owing to their impressive generalization capability, these models can be used directly for zero/few-shot classification tasks~\cite{gpt3, wu2023exploring} or indirectly to generate pseudo labels to train custom models~\cite{trueteacher, zhang2023pieclass}. They also serve as base models that can be fine-tuned on specific classification tasks~\cite{fingpt, kheiri2023sentimentgpt, li2023label}, achieving performance that matches/surpasses other architectures. Companies like OpenAI even provide APIs to fine-tune LLMs on custom data~\cite{openai_finetune}.

\begin{figure}[t]
  \centering
  \includegraphics[width=\columnwidth]{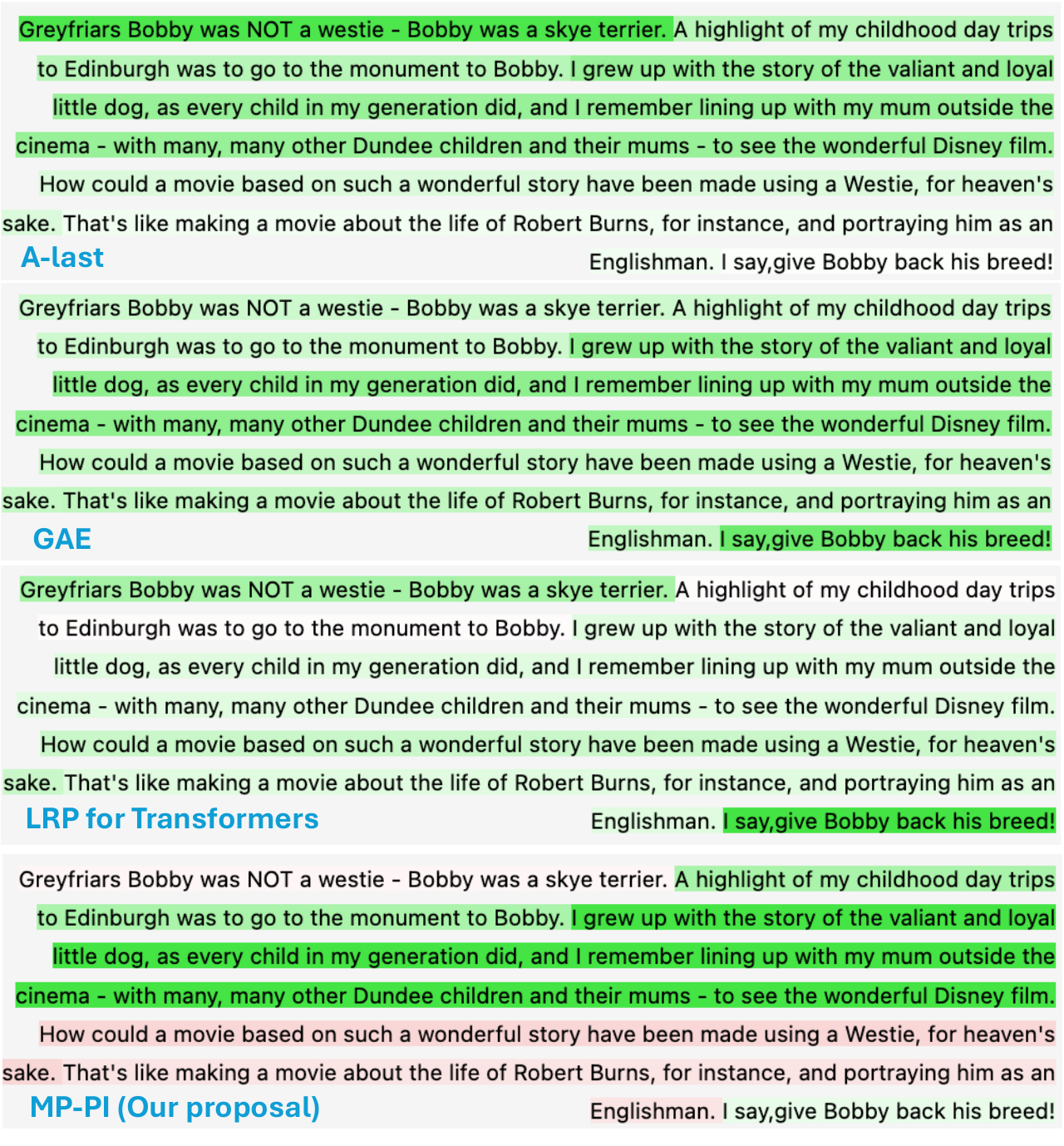}
  \vspace{-0.3in}
  \caption{Comparing the attributions produced by MP-PI with prior works on a misclassified movie review from the IMDB dataset. Only MP-PI manages to correctly identify negative sentences.}
  \label{fig:example}
\end{figure}

With the growing adoption of these models in critical applications such as healthcare and finance~\cite{bloomberg}, there is a strong need to provide accurate explanations to improve trust in the model's predictions. Input attribution is a form of explanation that addresses this need by highlighting input features that support/oppose the prediction of the model. This can be used to easily evaluate the correctness of the model's prediction, debug model performance~\cite{anders2022finding}, perform feature selection~\cite{zacharias2022designing}, and also to improve model performance by guiding the model to focus on the relevant parts of the input~\cite{krishna2023post}. While there are several prior works on generating input attributions using input perturbations~\cite{lundberg2017unified}, relevance propagation~\cite{ali2022xai}, attention scores~\cite{aflow}, or gradients~\cite{int_grad}, they are either expensive or yield low-quality attributions that do not accurately reflect the model's behavior (see Fig~\ref{fig:example} for an example). The goal of our work is to design a framework that provides high-quality explanations for decoder-only Transformer models by leveraging the unique properties of this architecture. 

%\textbf{Leveraging masked attention for explanations.} 
To this end, we start by observing that decoder-only models that are trained autoregressively use the masked self-attention mechanism. This mechanism enforces the property that the prediction of the model at any position only depends on the tokens seen at or before that position. Our key insight is that this property can be exploited to obtain the model's predictions on perturbed versions of the input, which can then be used to compute token/word/sentence-level attributions. To illustrate, consider the example in Fig.~\ref{fig:overview}. The input sequence $\{t_1, t_2, ..., t_n\}$ when passed through the decoder-only model produces the predictions $\{\vec{p}_1, \vec{p}_2, ..., \vec{p}_n\}$. Due to the causal attention mechanism, the prediction at the $i$-th position $\vec{p}_i$ only depends on tokens $\{t_1, t_2,...t_i\}$, which appear at or before the $i$-th position. As such, $p_i$ can be treated as the model's prediction on a perturbed/masked version of the input, where only the tokens/features $\{t_1, t_2,...t_i\}$ are active and the remaining tokens  $\{t_{i+1}, t_{i+2},...t_n\}$ are masked out. Thus, simply by computing the intermediate predictions, we can obtain the model's prediction on $n$ perturbed versions of the input, for almost no extra cost!

We develop a framework called $\emph{progressive inference}$ to produce highly-faithful explanations using the intermediate predictions from decoder-only models. We propose two methods that can be used under different compute budgets to explain decoder-only sequence classification models.

\emph{1. Single-Pass Progressive Inference (SP-PI):} SP-PI computes attributions over input features by taking the difference between consecutive intermediate predictions. This technique does not require additional forward passes and incurs negligible computational overheads to compute intermediate predictions. Despite its simplicity, we show through our experiments that it yields attributions that are on par or better than prior explainable AI (XAI) techniques that have a comparable amount of computational overhead.

\emph{2. Multi-Pass Progressive Inference (MP-PI):} A key limitation of SP-PI is that it does not have any control over the distribution of the masked inputs. E.g. in Fig.~\ref{fig:overview}, SP-PI only provides predictions associated with masked inputs, where the set of active features are of the form $\{t_1, t_2,...,t_i\}$. It is not possible to get the prediction on a masked input with an arbitrarily set of active features like $\{t_1, t_4, t_9\}$. MP-PI solves this problem by performing multiple inference passes with several randomly sampled masked versions of the input. Each inference pass yields intermediate predictions corresponding to a new set of perturbed inputs. To compute attributions with these predictions, we make a connection to Kernel SHAP~\cite{lundberg2017unified} by noting that intermediate predictions can be used to solve a weighted regression problem to compute input attributions. These attributions approximate SHAP values if the intermediate masks follow the Shapley distribution~\cite{shapley_distribution}. To this end, we design an optimization problem to find a probability distribution for sampling input masks, which results in the intermediate masks following the Shapley distribution. Owing to its principled formulation, MP-PI provides SHAP-like attributions that more accurately reflect the model's behavior compared to SP-PI and prior works.

In summary, we make the following key contributions:
\begin{enumerate}[leftmargin=0.5cm, noitemsep, topsep=0pt]
    \item We propose the \emph{Progressive Inference} framework that interprets the intermediate predictions of a decoder-only model as the approximate prediction of the model on masked versions of the input.
    \item We develop \emph{Single-Pass Progressive inference} -- a simple method that uses intermediate predictions to produce input attributions that explain the predictions of decoder-only models with negligible computational overheads.
    \item We propose \emph{Multi-Pass Progressive inference}-- a more complex explanation method, which uses multiple inference passes with masked versions of the input. A key part of our method is developing an optimization procedure to find a probability distribution for sampling input masks that results in SHAP-like attributions.
    \item We perform extensive perturbation studies to evaluate the quality of attributions. We show that our methods produce significantly better attributions compared to a wide suite of prior works, across different models (GPT-2, Llama-2 7b~\cite{llama2}), fine-tuned on a 7 different text classification tasks (sentiment classsification, natural language inference and news categorization).
\end{enumerate}

\section{Background and Related Work}
There is a rich body of prior works that have been proposed to compute feature attributions for DNNs. Additionally, several XAI methods have been developed specifically in the context of Transformer models. In this section, we start by formally defining the objective of input attribution techniques. We then provide an overview of these prior works. Through experimental evaluations, we show that our proposed SP-PI and MP-PI techniques provide higher quality explanations compared to these prior works.

\subsection{Problem Formulation}
Consider a model $f: \mathbb{R}^n \rightarrow \mathbb{R}^k$ that is trained to perform a $k$-class classification task. Let $N=\{1, 2, .., n\}$ denote the set of feature indices and $\vec{x}=[t_1, t_2, ...t_n]$ denote the input vector, where $t_i$ represents the $i^{th}$ feature/token. The goal of input-attributions techniques is to compute feature-level attributions $\vec{\phi} =[\phi_1, \phi_2, ..., \phi_n]$ that reflects the influence of each feature on the prediction of the model. These attributions can either be computed for each token or groups of tokens (representing words/sentences).

\subsection{Perturbation-based Methods}
Perturbation-based methods are based on the idea that the importance of input features can be measured by examining how the prediction of the model changes for different perturbed versions of the input. The most principled formulation of this idea is the SHAP framework~\cite{strumbelj2010efficient, lundberg2017unified} that computes input attributions by using a game-theoretic approach that views input features as players and the prediction of the model as the outcome in a collaborative game. The attribution $\phi_i$ for the $i^{th}$ feature can be computed by taking a weighted average of the marginal contributions of the $i^{th}$ feature, when added to different coalitions of features $S$, as shown below
\begin{equation}\label{eq:shap}
\phi_i=\sum_{S\subseteq N \setminus \{i\}}\frac{|S|!(n-|S|-1)!}{n!} \left[f(x_{S \cup \{i\}}) - f(x_S)\right].
\end{equation}
The feature attributions computed this way are called SHAP values~\cite{shapley} and have been shown to satisfy several desirable axiomatic properties like local-accuracy, missingness, and consistency~\cite{young1985monotonic, lundberg2017unified}. Since the number of terms in the SHAP equation grows exponentially with the number of input features, computing it exactly is intractable when there are a large number of features in the input. To mitigate this issue, sampling-based methods such as Sampling SHAP and Kernel SHAP~\cite{lundberg2017unified} have been proposed to compute approximate SHAP values in a tractable way. Sampling SHAP simply evaluates a subset of the terms in Eqn.~\ref{eq:shap}, while Kernel SHAP uses the idea that SHAP values can be viewed as a solution to the following weighted linear regression problem (with weights $w(S)$):
\begin{align}\label{eq:kernel_shap}
\{\phi_i\} &= \argmin\limits_{\phi_1,..\phi_n} \sum\limits_{S\subseteq N}w(S)\Big(f(t_S)-g(S)\Big)^2\\
&\text{where, } g(S) = \phi_0 + \sum\limits_{i\in S}\phi_i
\end{align}
Our proposed methods \emph{SP-PI} and \emph{MP-PI} also fall under the category of perturbation based methods, as they both leverage the model's prediction on perturbed versions of the input to compute feature attributions. Furthermore, \emph{MP-PI} uses a connection with Kernel SHAP to compute SHAP-like attributions more efficiently compared to Kernel SHAP.

\subsection{Gradients, Activations, and Propagation Rules}
Several methods to compute attributions have been proposed by using some combination of gradients, activations, and propagation rules to compute input attributions. Gradient $\times$ Input~\cite{shrikumar2016not} is one such method that uses a product of gradients and inputs to compute attributions. Integrated gradients~\cite{sundararajan2017axiomatic} generalizes this approach by first computing the average gradient along the straightline path between a baseline input $\vec{t_b}$ and the actual input $\vec{t}$. This average gradient is multiplied with difference in the input and baseline to compute the attribution. Layer-wise Relevance Propagation (LRP)~\cite{bach2015pixel} is another XAI method for DNNs that is based on the idea that the relevance score of the output neurons of a layer can be redistributed to the input neurons using propagation-rules. LRP recursively applies propagation rules, starting from the last layer, going backwards, until the relevance-scores for the input features (i.e. attributions $\vec{\phi}$) can be computed. DeepLIFT~\cite{shrikumar2017learning} is a generalization of LRP that uses a baseline input as reference to compute relevance scores.

% Progressive Inference (PI) treats intermediate predictions $\vec{p_i}$ as the predictions of the model on masked versions of the input $\vec{x_i}$.

\subsection{Methods for Transformers}
Recent works have developed XAI techniques, specifically to explain Transformer models. Transformer models are based on the attention mechanism, where attention scores are used in each Transformer block to produce the output by taking a weighted average over the input tokens. Several works have repurposed the attention scores to produce input attributions. Among these methods are attention-last~\cite{a_last}, which directly uses the last-layer attention. Attention-flow and attention-rollout~\cite{a_flow} compute attributions by capturing information flow using the attention weights. Generic attention-model explainability (GAE)~\cite{gae} uses a combination of attention and gradient maps to generate relevance maps. LRP for Transformers~\cite{xai_for_transformers} is a recent technique that adapts the LRP technique to the Transformer architecture by changing the way relevance scores are propagated through the layers.  

\begin{figure}[t]
  \centering
  \includegraphics[width=\columnwidth]{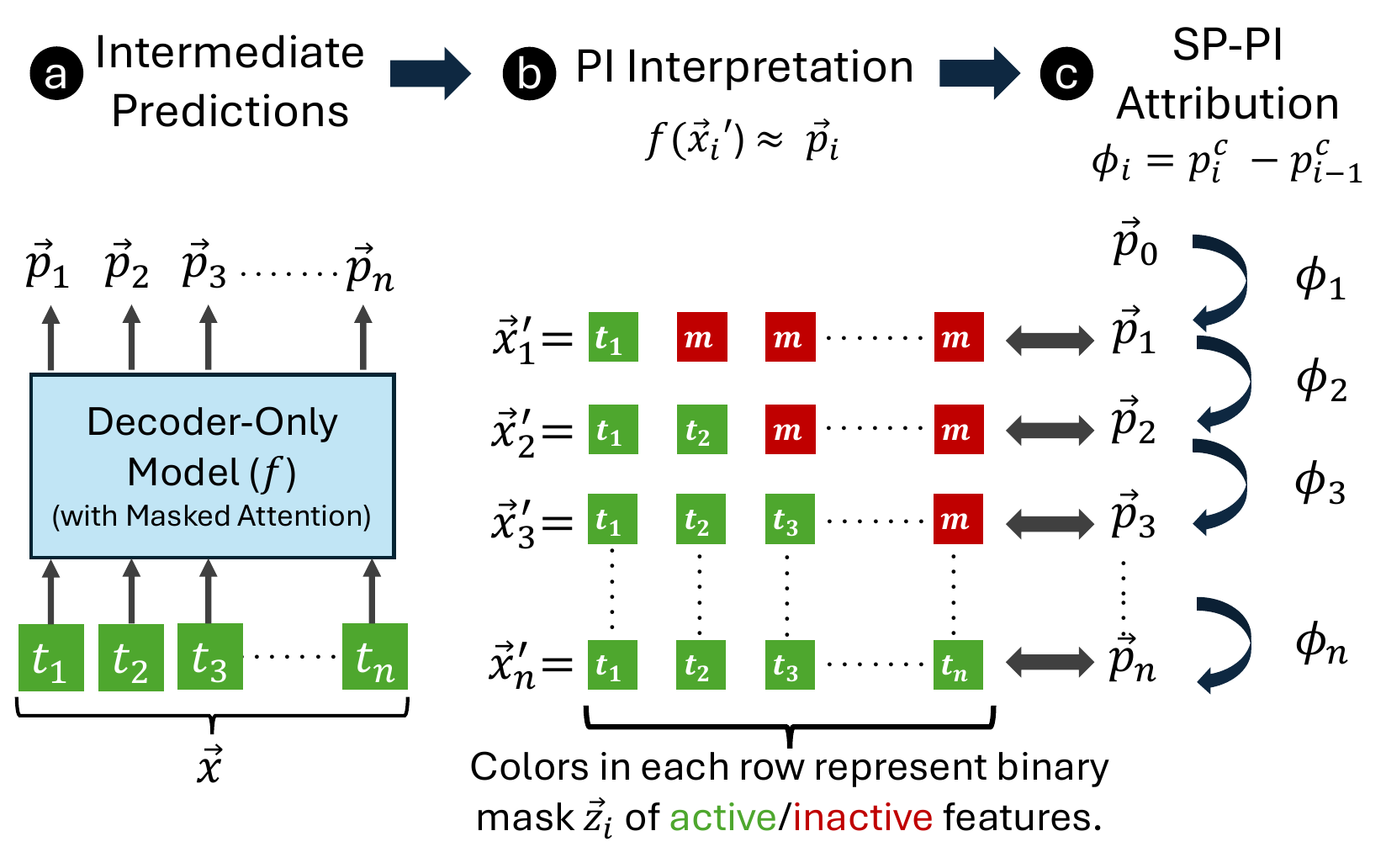}
  \vspace{-0.2in}
  \caption{\circled{a} SP-PI uses the original input $\vec{x}$ to produce intermediate predictions $\{\vec{p_i}\}$. \circled{b} The PI framework treats these intermediate predictions as approximations of the model's prediction on the corresponding masked versions of the inputs: $\vec{p_i}\approx f(\vec{x'_i})$. \circled{c} SP-PI takes the difference in the intermediate predictions to compute feature-level attributions $\{\phi_i\}$.}
  \label{fig:sppi}
\end{figure}
\section{Progressive Inference}
We propose the Progressive Inference (PI) framework for computing input attributions to explain the predictions of decoder-only models.
%PI is a perturbation-based method that exploits the masked-attention mechanism of decoder-only Transformer models to efficiently approximate the prediction of the model on perturbed versions of the input. 
PI exploits the key observation that the intermediate predictions of a decoder-only model only depends on the tokens that appear at or before that position. We use this observation to interpret intermediate predictions as representing the prediction of the model on masked versions of the input.

To explain, consider Fig.~\ref{fig:sppi}a, where the input $\vec{x}=[t_1, t_2,..,t_n]$ is passed through the model $f$ to produce the intermediate predictions $\{\vec{p}_1, \vec{p}_2,..., \vec{p}_n\}$. Due to the causal attention mechanism, we can intuitively view $\vec{p_1}, \vec{p_2},...,\vec{p_n}$ as representing the predictions of the model on the masked inputs $[t_1, m,...,m], [t_1, t_2,...,m],...,[t_1, t_2,...,t_n]$ respectively. More formally, we interpret $\vec{p_i}$ as an approximation of the model's prediction on perturbed/masked versions of the original input as shown in Eqn.~\ref{eq:1}.\footnote{The approximation error in Eqn.~\ref{eq:1} can vary with prediction position, length of the input and the model being used. Regardless, this is a useful interpretation that lets us connect progressive inference with other perturbation techniques like SHAP.}
\begin{align}
    \vec{p_i} &\approx f(\vec{x}'_i), \label{eq:1} \\
    \text{where } \vec{x}'_i = h_{\vec{x}}(\vec{z}_i) &= \vec{z}_i \odot \vec{x} + (1-\vec{z}_i) \odot m.  \label{eq:2}
\end{align}
Here, $\vec{z}_i$ is a binary mask vector which indicates the features that are active in the perturbed input $\vec{x'_i}$ as shown in Fig.~\ref{fig:sppi}b. To reflect the causal attention mechanism, we set $\vec{z}_i$ to be the $i^{th}$ row of a $n\times n$ lower triangular matrix of ones $\mathcal{L}_1$.
%Here, $\vec{z_i}=[z^1_i, z^2_i, ..z^n_i]$ is a binary mask vector such that $z^j_i = [1 \text{ for } j \leq i, \text{ and } 0 \text{ otherwise}]$. This binary mask 
$h_{\vec{x}}: \mathbb{Z} \rightarrow \mathbb{X}$ is a masking function that maps the binary mask to the masked input as defined by Eqn~\ref{eq:2}. $m$ denotes the mask token that is used to replace inactive tokens. 

Using the above interpretation, with a single forward pass of the model, we can obtain the prediction of the model on up to $n$ perturbed inputs: $\{(\vec{x}'_i, \vec{p}_i)\}$. 
%Note that, obtaining these predictions without PI would require $n$ forward passes.
We can use this set of  $(\vec{x'_i}, \vec{p_i})$ pairs  to compute input attributions that explain the prediction of the model. 

We describe two methods to compute input attributions.
%using the set of $(\vec{x}'_i, \vec{p}_i)$ pairs. 
We start by describing \emph{Single-Pass Progressive Inference (SP-PI)}--a simple low-cost technique to compute attributions that only requires a single forward inference pass through the network. We then propose a more complex technique called \emph{Multi-Pass Progressive Inference} (MP-PI), which uses the intermediate predictions collected from multiple inference passes using masked versions of the inputs. MP-PI leverages a connection with Kernel SHAP to compute higher quality attributions.

\subsection{Single-Pass Progressive Inference}
SP-PI requires a single forward-pass with the original input $\vec{x}$. Let $\vec{p}_i = [p_i^1, p_i^2, .. p_i^k]$ denote the logit-vector associated with the $i^{th}$ intermediate prediction. To explain the model's prediction on class $c$, SP-PI computes attribution for the $i^{th}$ feature by taking the difference between successive intermediate predictions as follows:
%We compute the attribution for the $i^{th}$ token ($\phi_i$) by taking the difference between successive intermediate predictions as follows:
\begin{align}
    \phi_i = p_i^c - p_{i-1}^c \label{eq:3}
\end{align}
%Here, $\phi_i$ explains the model's prediction on class $c$. 
We note that the attribution $\phi_i$ is quite simply the \emph{change in the model's prediction after seeing the $i^{th}$ feature}. More formally, the attribution $\phi_i$ can be viewed as the marginal change in the prediction of the model, when the $i^{th}$ feature is added to the coalition of features $\bar{S}_{i-1} = \{1, 2, ..i-1\} $ that came before it. This can be seen more clearly by using Eqn.~\ref{eq:1},~\ref{eq:2} to rewrite Eqn.~\ref{eq:3} as follows:
\begin{align}\label{eq:4}
    \phi_i &\approx f^c(\vec{x}'_i) - f^c(\vec{x}'_{i-1}),\\
    \phi_i &\approx  f^c(h_{\vec{x}}(\vec{z}_{\bar{S}_{i-1}\cup \{i\}})) - f^c(h_{\vec{x}}(\vec{z}_{\bar{S}_{i-1}})).
\end{align}
Here, $S$ denotes a set of active features and $\vec{z_S}$ denote the corresponding binary mask vector such that $z^j_S = [1 \text{ for } j \in S, \text{ and } 0 \text{ otherwise}]$.
\begin{figure*}[t]
  \centering
  \includegraphics[width=1.0\textwidth]{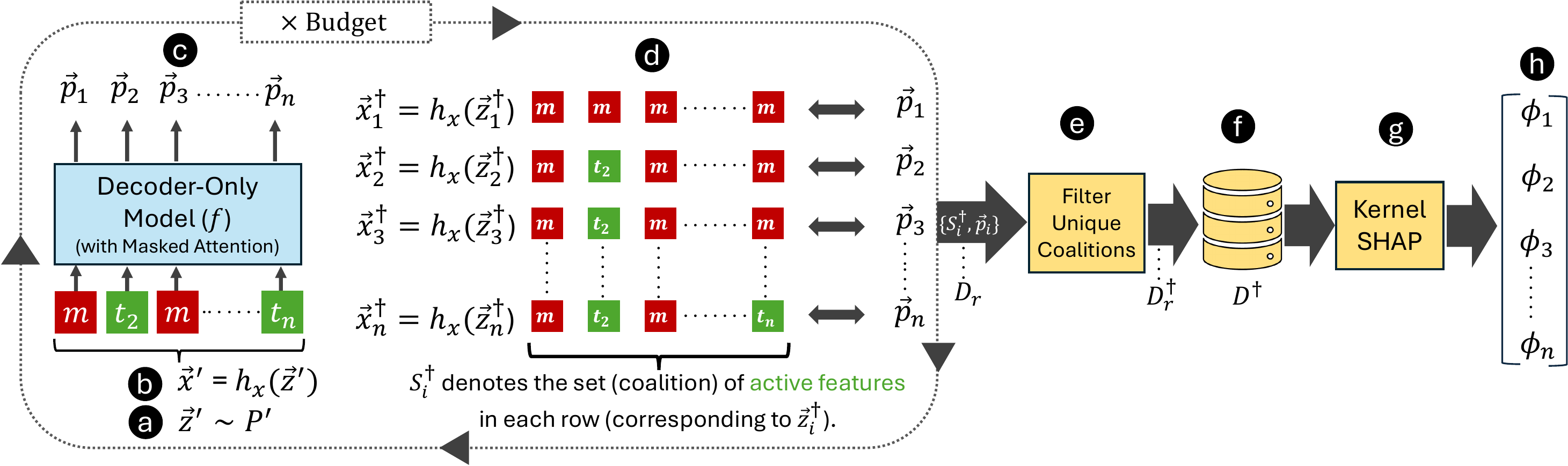}
  \vspace{-0.2in}
  \caption{ MP-PI runs progressive inference multiple times with different masked versions of the input. It starts by \circled{a} sampling a binary mask $\vec{z}'$ to create a \circled{b} masked input $\vec{x}'$. PI interprets the \circled{c} intermediate predictions $\{\vec{p}_i\}$ generated from $\vec{x}'$ as predictions of the model on \circled{d} different perturbed versions of the input $\{\vec{x}'_i=h_{\vec{x}}(\vec{z}'_i)\}$. \circled{e} The set of (coalition, prediction) pairs $(S_i, \vec{p}_i)$ are filtered to remove repeated coalitions and \circled{f} added to the dataset $D$. Finally, we use \circled{g} Kernel SHAP on $D$ to produce the \circled{h} input attributions $\{\phi_i\}$.}
  \label{fig:mppi}
  \vspace{-0.1in}
\end{figure*}

\textbf{Connection to SHAP Values.}
Both SHAP and SP-PI compute attributions by evaluating the change in the model's prediction by adding a feature to a coalition of features. SHAP computes feature attribution by considering the weighted average of a feature's marginal contribution across multiple coalitions (Eqn.~\ref{eq:shap}). In contrast, SP-PI computes attribution by only considering a single coalition (Eqn.~\ref{eq:4}). While both SP-PI and SHAP satisfy desirable axiomatic properties like \emph{local accuracy} (see Proposition 1 in Appendix~\ref{app:proof} for proof), the quality of attributions computed with SP-PI falls short of SHAP values as SP-PI only considers a single coalition.

\subsection{Multi-Pass Progressive Inference}

 %A key shortcoming of SP-PI is that, to compute $\phi_i$, it only considers the marginal change in the output of the function when adding the $i^{th}$ feature to a \emph{single coalition} of the form $S_{i-1} = \{j\}_{j=1}^{i-1}$. In contrast, SHAP computes $\phi_i$ by considering the influence of adding the $i^{th}$ feature to multiple possible coalitions of features. 
A key limitation of SP-PI is that, to compute $\phi_i$, it considers a single coalition of features of the form $\bar{S}_{i-1} = \{1, 2, 3,...,i-1\}$ (i.e. the set of all features that appear before the $i^{th}$ feature). 
%the coalition of features considered to compute the marginal contribution of the $i^{th}$ feature $t_i$ is always of the form  $S_{i-1} = \{1, 2, 3,...,i-1\}$ (i.e. the set of all features that appear before $t_i$). 
This prevents us from evaluating the marginal contribution on arbitrary subsets of features as is done with SHAP values. To bridge this gap, we propose multi-pass progressive inference (MP-PI). 

%\textbf{Overview of MP-PI:} 
\subsubsection{Overview}
MP-PI performs multiple rounds of progressive inference, each time with a different masked version of the input, allowing us to sample a more diverse coalition of features. Fig.~\ref{fig:mppi} provides a visual depiction of MP-PI. In each round, we start by sampling a binary mask $\vec{z}'$ from a pre-defined masking distribution $P'$ (Fig.~\ref{fig:mppi}a). We use $\vec{z}'$ to obtain a masked version of the input $\vec{x}' =h_{\vec{x}}(\vec{z}')$ (Fig.~\ref{fig:mppi}b). We perform inference on this masked input to obtain the set of intermediate predictions $\{\vec{p}_i\}$ (Fig.~\ref{fig:mppi}c). Using the PI interpretation (Fig.~\ref{fig:mppi}d), we have
\begin{align}
    \vec{p}_i&\approx f(\vec{x}_i^\dagger),\\
    \text{where }\vec{x}_i^\dagger&=h_{\vec{x}}(\vec{z}_i^\dagger), \vec{z}_i^\dagger=\vec{z}'\odot\vec{z}_i.
\end{align}
Here, $\vec{x}_i^\dagger$ denotes the perturbed input corresponding to $\vec{p}_i$, $\vec{z}_i^\dagger$ is the binary mask applied to $\vec{x}$ to produce $\vec{x}_i^\dagger$. $\vec{z}_i^\dagger$ can be expressed as the Hadamard product of $\vec{z}'$ (the masking vector used to produce $\vec{x}'$) and $\vec{z}_i$ ($i^{th}$ row of $\mathcal{L}_1$ i.e. the lower triangular matrix of ones). We use $S_i^\dagger$ to denote the set (i.e. coalition) of active features in $\vec{z}_i^\dagger$. Let $D_r$ represent the set $\{S^\dagger_i, \vec{p_i}\}$ collected in the $r^{th}$ round. Note that $D_r$ can have redundant coalitions (e.g. $S_2$ and $S_3$ in Fig.~\ref{fig:mppi} have the same set of features). We filter $D_r$ to only retain unique coalitions to create $D_r^\dagger$ (Fig.~\ref{fig:mppi}e). The $D_r^\dagger$ from each round are combined to construct the dataset $D^\dagger$ (Fig.~\ref{fig:mppi}f). We then use Kernel SHAP (Fig.~\ref{fig:mppi}g) with this dataset to compute the feature attributions $\{\phi_i\}$ (Fig.~\ref{fig:mppi}h). This procedure is also described more formally in Algorithm~\ref{alg:mppi}

\begin{algorithm} 
    \caption{Multi-pass progressive Inference}
    \label{alg:mppi}
\begin{algorithmic}
    %\Procedure{MPPI}{$f$, $\vec{x}$, $\mathcal{B}$, $\mathcal{P}_{\mathcal{M}}$}
    \STATE {\bfseries Inputs:} model $f$, input vector $\vec{x}$, budget $\mathcal{B}$, mask sampling distribution $P'$
    \STATE $n \gets |x|, \{\vec{z}_i \gets \mathcal{L}_1[i]\}, D^\dagger\gets\{\}$
    \FOR{$r \gets 1$ to $\mathcal{B}$}
    \STATE $\vec{z}' \sim P'$
    \STATE $\vec{x}^\dagger \gets h_{\vec{x}}(\vec{z}')$
    \STATE $\{\vec{p}_i\} \gets f_{inter}(\vec{x}^\dagger)$
    \STATE $\{\vec{z}_i^\dagger \gets  \vec{z}' \odot \vec{z}_i\}$
    \STATE $\{S_i^\dagger\gets \mathbb{S}(\vec{z}_i^\dagger)\}$ 
    \STATE $D_r \gets\{S_i^\dagger, \vec{p}_i\}$
    \STATE $D_r^\dagger \gets filter\_unique\_coalitions(D_r)$
    \STATE $D^\dagger \gets D\cup D'_r$   
    \ENDFOR
    \STATE $\{\phi_i\} = KernelSHAP(D^\dagger)$
    %\STATE {\bfseries Return }$\{\phi_i\}$
    %\EndProcedure
\end{algorithmic}
\end{algorithm}

\subsubsection{Using Kernel SHAP to Compute $\phi_i$}
%KernelSHAP computes attributions by framing it as the solution to a linear regression problem. 
Kernel SHAP starts by defining a linear model $g(S) = \phi_0 + \sum\limits_{i\in S}\phi_i$, where $S\subseteq N$ denotes a coalition of input features. The coefficients $\{\phi_i\}$ are optimized using the dataset $D^\dagger$ by solving the weighted linear regression problem in Eqn.~\ref{eq:ks}.
\begin{align}\label{eq:ks}
    \{\phi^*_i\} = \argmin\limits_{\phi_1,..\phi_n} \sum\limits_{(S_i^\dagger, \vec{p}_i)\in D^\dagger}w(S^\dagger_i)\Big(p_i^c-g(S^\dagger_i)\Big)^2.
\end{align}
If the coalitions in $D^\dagger$ are sampled independently and their distribution (denoted by $P^D$) follows the \emph{Shapley  distribution} $P^*$, then the solution $\{\phi^*_i\}$, obtained by optimizing Eqn~\ref{eq:ks} with uniform weights $w(S_i^\dagger)$, represent the SHAP values. Unfortunately, the samples in $D^\dagger$ are not independently sampled. However, we have the ability to control $P^D$  by carefully selecting the distribution of masks $P'$, which is used to generate the perturbed inputs $\vec{x}'$ (i.e. the masked input to the model in Fig.~\ref{fig:mppi}a). Thus, for $\{\phi^*_i\}$  to approximate SHAP values, \textbf{we need to find an optimal $P'$ that results in $P^D$ following the Shapley distribution $P^*$.}

\subsubsection{Optimizing $P'$}
%Before describing the optimization procedure to find the optimal $P'$, 
We start by introducing some notation to express $P^*$ (Shapley distribution) and $P'$ (input masking distribution). We then establish a connection between $P'$ and $P^D$ (distribution of intermediate coalitions). Finally, we formulate an optimization procedure to find the $P'$ that minimizes the distance between $P^D$ and $P^*$.

\textbf{Notations for $P^*$:} The Shapley distribution can be expressed in a vector form as $[P^*_1, P^*_2,...,P^*_{n-1}]$, where $P^*_i = \frac{1}{Ci(n-i)}$ denotes the probability of sampling a coalition of size $i$. Here, $C=\sum_i\frac{1}{i(n-i)}$ is the normalization constant that ensures that $\sum_i P^*_i = 1$. Alternatively, $P^*$ can also be expressed as an $(n-1)\times n$ matrix, where each entry of the matrix $P^*_{ij}$ indicates the probability of sampling coalitions of size $i$, where $j$ is the last active feature. More formally, we can write this as
\begin{align}\label{eq:pij}
    P^*_{ij} =\text{Pr}\left( S_{ij} : |S_{ij}| = i, j \in S_{ij}, \forall k\in N/S_{ij}, k > j\right).
\end{align}
Note that $S_{ij}$ does not refer to any single coalition of features as there are multiple coalitions that could satisfy the conditions for $S_{ij}$ in Eqn.~\ref{eq:pij}. We can express $P^*_{ij}$ in terms of $P^*_i$ as follows (see Proposition 2 in Appendix~\ref{app:proof} for proof):
\begin{align}
P^*_{ij} = \begin{cases}
P^*_i \binom{j-1}{i-1}/\binom{n}{i} & \text{if } j \geq i \\
0 & \text{otherwise.}
\end{cases}
\end{align}
\textbf{Notations for $P'$:} Similarly, we can express the masking distribution $P'$ as a $(n-1)\times n$ matrix consisting of entries $P'_{ij}$ that indicate the probability of sampling coalitions of the form $S_{ij}$, where $|S_{ij}|=i$ and $j$ is the last active feature. 

\begin{table}[b]
\caption{Details of datasets, models and attribution types used in our experiments.}
\centering 
\scalebox{0.8}{
\begin{tabular}{l|lllllll}
\toprule
Dataset (n. classes) & Model (size) & Acc.$\%$ & Source & Attr.\\
\midrule
IMDB (2) & GPT-2 (124M) & 94.06 & FFT (HF) & Sent.\\
SST-2 (2) & GPT-2 (355M) & 92 & FFT (HF)& Word\\
AG-News (4) & Llama-2 (7B) &94.96 & PEFT& Word\\
Twitter-Fin (3) & Llama-2 (7B) & 91.08 & PEFT& Word \\
Twitter-Sentiment (3) & GPT-2 (124M) & 68.18& FFT & Word\\
Twitter-Emotion (4) & GPT-2 (124M) & 80.29 & FFT & Word\\
TrueTeacher NLI (2) & GPT-2 (1.5B) & 86.21 & PEFT& Sent.\\
\bottomrule
\end{tabular}
}
\label{tab:datasets}
\end{table}

\def\rot{0}
\begin{table*}[t]
\caption{AUC $(\uparrow)$ for the activation study comparing different XAI methods. A higher AUC indicates better performance. \emph{Cost} indicates the compute (normalized to a single inference pass) required to generate attributions for each method. For each dataset, the best AUC among all methods is marked in \textbf{bold} and among methods with cost $\leq 1\times$ is marked with an \underline{underline}. SP-PI and MP-PI provide the best explanations for most datasets in their respective cost categories.}
\centering 
\scalebox{0.8}{
\begin{tabular}{ll|lllllll}
\toprule
Method       &Cost ($\times$)& \rotatebox{\rot}{IMDB} & \rotatebox{\rot}{SST-2} & \rotatebox{\rot}{AG-news} &   \rotatebox{\rot}{Twitter-Fin} & \rotatebox{\rot}{Twitter-Sen} & \rotatebox{\rot}{Twitter-Emo} & \rotatebox{\rot}{TrueTeacher} \\
\midrule
Random          &0& 0.855& 0.756&  0.763& 0.763& 0.583& 0.624& 0.699\\
A-Last          &0& 0.873& 0.754&  \underline{0.855}&  0.808&   0.627& 0.715&0.781\\
\textbf{SP-PI}  &0& \underline{0.951}& 0.84&  0.817&  \underline{0.814}& \underline{0.747}& \underline{0.869}& \underline{0.879}\\
GAE             &1& 0.916& \underline{0.863}&  0.782&  0.795&  0.687& 0.792& 0.829\\
Inp X Grad      &1& 0.903& 0.811&  0.772&  0.806&  0.646& 0.779& 0.833\\
LRP for Trfm.   &1& 0.901& 0.826&  0.776&  0.811&  0.648& 0.781& 0.835\\
\midrule
Int. Grad       &$2n$& 0.9& 0.877&  0.755&  0.791&  0.739& 0.832& 0.826\\
Kernel SHAP     &$2n$& 0.959& 0.899&  0.778&  0.821&  0.816&  0.871& 0.811\\
\textbf{MP-PI}  &$2n$& \textbf{0.97}& \textbf{0.946}&  \textbf{0.867}&  \textbf{0.847}&  \textbf{0.887}& \textbf{0.929}& \textbf{0.921}\\
\bottomrule
\end{tabular}
}
\label{tab:auc_act}
\end{table*}
\textbf{Connecting $P'$ and $P^D$:} In the PI framework, predictions on an input coalition $S'_{ij}$ (representing the input $\vec{x}'$ in Fig.~\ref{fig:mppi}b), yields additional predictions for coalitions of the form $\{S^\dagger_{kl}\}_{k=1}^i$ i.e. coalitions of sizes $1,2,..,i$ (represented by $D^\dagger_r$ in Fig.~\ref{fig:mppi}e). We can view the distribution of these additional coalitions $S^\dagger_{kl}$ as being conditioned on $S'_{ij}$. Assuming $i,j,k,l\in N$, this conditional distribution is given by (see Proposition 3 in Appendix~\ref{app:proof} for proof):
\begin{align}\label{eq:ijkl}
P^\dagger_{kl|ij} = \begin{cases}
\binom{l-1}{k-1}\binom{j-l}{i-k}/(\binom{j-1}{i-1}i) & \text{if } k \leq i, l\leq j, j\geq i \\
0 & \text{otherwise.}
\end{cases}
\end{align}
There are $n(n-1)$ values for $i,j$ and $k,l$. Thus, $P^\dagger_{kl|ij}$ can be written as a $n(n-1) \times n(n-1)$ matrix. We can use this conditional distribution matrix to express $P^D$ in terms of $P'$ as follows
\begin{align}\label{eq:pcond}
    \vec{P}^D = \vec{P}' P^\dagger_{kl|ij}.
\end{align}
Here, $\vec{P}^D$ and $\vec{P}'$ are the vectorized representation of the matrices $P^D$ and $P'$. Note that our goal is to optimize $P'$ to minimize the distance between $P^D$ and $P^*$. We can do so by solving the following optimization problem:
\begin{align}\label{eq:opt}
    P' = \argmin_{P'}{|\vec{P}' P^\dagger_{kl|ij} - \vec{P}^*|} \text{ s.t. } P'_{ij}\geq 0.
\end{align}
Note that the $P'$ obtained from Eqn.~\ref{eq:opt} may not result in $P^D$ exactly matching $P^*$. We remedy this issue by setting $w(S_{ij})=P^*/P^D$ in Eqn.~\ref{eq:kernel_shap} when computing the attributions with Kernel SHAP.

\subsubsection{Maximizing the Number of Samples}
While the procedure described thus far is sufficient to find SHAP-like attributions, we can perform one final optimization to maximize the number of coalitions that we obtain when running MP-PI. We start by noting that the intermediate coalitions $\{S^\dagger\}$ obtained by an input coalitions $S'_{ij}$ is a subset of the intermediate coalitions obtained by the input coalition $S^+_{ij}=S'_{ij}\cup \{j+1, j+2,...,n\}$, where $n$ is the total number of input features. To illustrate, consider the input coalition $S'=\{1,3,4\}$, with $n=6$. By running PI with $S'$, we obtain $3$ unique coalitions $\{S^\dagger\}$: $\{\{1\}, \{1,3\}, \{1,3,4\}\}$. Instead, if we modify $S'$ to include $\{5,6\}$ i.e. $S^+=\{1,3,4,5,6\}$, we get the following unique intermediate coalitions with PI: $\{S^\dagger\}:$ $\{\{1\}, \{1,3\}, \{1,3,4\}, \{1,3,4,5\}, \{1,3,4,5,6\}\}$. Note that this contains all the coalitions provided by $S'$, and two extra coalitions: $\{1,3,4,5\}$ and $\{1,3,4,5,6\}$. 

%In general, using $S^+$ is always better than $S'$, as it provides additional samples for no extra cost.

To maximize the number of coalitions, we use $S_{ij}^+$ in MP-PI instead of $S_{ij}'$. Due to this modification, the conditional distribution in Eqn.~\ref{eq:ijkl} changes to the following
\begin{align}\label{eqn:ijkl+}
P^\dagger_{kl|ij} = \begin{cases}
\binom{l-1}{k-1}\binom{j-l}{i-k}/(\binom{j-1}{i-1}i') & \text{if } k < i, l<j, j \geq i  \\
1/i' & \text{if } l\geq j, k = i+l-j \\
0 & \text{otherwise.}
\end{cases}
\end{align}
Here, $i' = (i+n-j)$, which denotes the total number of active features in $S_{ij}^+$. We use the conditional distribution in Eqn.~\ref{eqn:ijkl+} instead of the one in Eqn.~\ref{eq:ijkl} to optimize $P'$.

\section{Experiments}

In this section, we compare the quality of attributions for our two proposed methods against a suite of prior works using a diverse set of classification tasks and models. We start by describing the experimental setup and then present the results showing the efficacy of our proposed techniques.

\begin{table*}[t]
\caption{AUC $(\downarrow)$ for the inverse activation study comparing different XAI methods. A lower AUC indicates better performance. For each dataset, the best AUC among all methods is marked in \textbf{bold} and among methods with cost $\leq 1\times$ is marked with an \underline{underline}. SP-PI and MP-PI provide the best explanations for most datasets in their respective cost categories.}
\centering 
\scalebox{0.8}{
\begin{tabular}{ll|lllllll}
\toprule
Method       &Cost($\times$)& \rotatebox{\rot}{IMDB} & \rotatebox{\rot}{SST-2} & \rotatebox{\rot}{AG-news} &   \rotatebox{\rot}{Twitter-Fin} & \rotatebox{\rot}{Twitter-Sen} & \rotatebox{\rot}{Twitter-Emo} & \rotatebox{\rot}{TrueTeacher} \\
\midrule
Random          &0& 0.863& 0.735&  0.758& 0.761& 0.595& 0.638&0.671\\
A-Last          &0& 0.839& 0.763&  0.62&  0.743&  0.557& 0.539& 0.556\\
\textbf{SP-PI}  &0& \underline{0.698}& 0.653&  0.692&  \underline{0.706}& \underline{0.429}& \underline{0.347}& \underline{0.455}\\
GAE             &1& 0.761& \underline{0.603}&  \underline{0.688}&  0.737& 0.513& 0.412& 0.476\\
Inp x Grad      &1& 0.793& 0.671&  0.697&  0.743& 0.529& 0.418&  0.476\\
LRP for Trfm.   &1& 0.795& 0.651&  0.698&  0.741& 0.519& 0.423&  0.468\\
\midrule
Int. Grad       &$2n$& 0.815& 0.582&  0.779&  0.761& 0.427& 0.361&  0.554\\
Kernel SHAP     &$2n$& 0.688& 0.533&  0.734&  0.709&  0.348& 0.337& 0.547\\
\textbf{MP-PI}  &$2n$& \textbf{0.613}& \textbf{0.431}&  \textbf{0.596}&  \textbf{0.684}&  \textbf{0.273}& \textbf{0.214}& \textbf{0.355}\\
\bottomrule
\end{tabular}
}
\label{tab:auc_act_inv}
\end{table*}

\subsection{Experimental Setup}

\textbf{Datasets and Models:} We pick a diverse set of sequence classification datasets (Table~\ref{tab:datasets}) and fine-tune decoder-only models of different sizes on these datasets. We generate explanations on the predictions of these models using different XAI methods and compare relative performance. For IMDB~\cite{imdb} and SST-2~\cite{sst2} datasets, we use models that are available on the HuggingFace repository.\footnote{IMDB: hipnologo/gpt2-imdb-finetune, SST-2: michelecafagna26 /gpt2-medium-finetuned-sst2-sentiment. The authors of these models are not affiliated with this paper.} For AG-News~\cite{agnews}, Twitter-Finance, Twitter-Sentiment~\cite{tweet_sent}, Twitter-Emotion~\cite{tweet_emo} and TrueTeacher~\cite{trueteacher} datasets, we fine-tune GPT-2~\cite{radford2018improving} or Llama-2~\cite{llama2} models with full fine tuning (FFT) for smaller models and parameter efficient fine tuning (PEFT) with LoRA~\cite{lora} for larger models. Additional details on training are provided in Appendix~\ref{app:training}

\textbf{Attribution Type:} We compute attributions for groups of tokens (instead of individual tokens) at either word or sentence level, as attributions at token-level may be too granular for a human reviewer to interpret. Note that Kernel SHAP, SP-PI and MP-PI can be straightforwardly adapted to compute word/sentence level attributions by considering groups of tokens (representing a word/sentence) as a single feature. For other methods, we aggregate token-level attributions to produce word/sentence level scores.

\textbf{Measuring the Quality of Attributions:} We randomly sample 500 examples from the test set of each dataset and compute attributions to explain the model's prediction on the true class $c$ with these examples. To quantify the quality of explanations, perform two studies with the attributions: %\emph{activation study} and a new study we propose called the \emph{inverse activation study}. 

%\begin{itemize}[noitemsep, leftmargin=*, topsep=0pt]
\emph{1. Activation study (AS):} AS~\cite{activation_task, xai_for_transformers} measures the ability of attributions to identify input features that increase the model's prediction on the chosen class. It works by sorting the input features in a descending order of attribution values $N_{AS} = argsort(\{-\phi_i\})$ (i.e. most positive to most negative). It then creates a fully masked version of the input and incrementally adds individual features from $N_{AS}$. Note that the model's prediction changes as new features are added. The probability corresponding to the correct class is plotted as a function of the number of features added and the Area Under the Curve (AUC) of this plot can be used to measure the quality of attribution.

\emph{2. Inverse Activation study (IAS):} In contrast to AS, the inverse activation study measures the ability of the attributions to identify features that reduce (negatively influence) the predictions of the model on the chosen class. Identifying such features is especially useful in the event of a misprediction--to override or debug the model's prediction (see Fig~\ref{fig:qualitative results1},~\ref{fig:qualitative results2} in Appendix~\ref{app:qualitative_results} for examples). IAS works by sorting features in an increasing order of attributions values $N_{IAS} = argsort(\{\phi_i\})$ (i.e. most negative to most positive). It then measures the AUC of the curve obtained by plotting the prediction of the model on the correct class $f^c(x')$. A lower AUC indicates better performance for IAS since features with negative influence are added first.
%\end{itemize}

\textbf{Prior Works:} Table~\ref{tab:auc_act} lists the representative set of prior works that are considered in our evaluations. This includes methods that use attention mechanism (GAE, A-Last), gradient based techniques (Inp x Grad, integrated gradients), relevance propagation methods(LRP for Transformers) and perturbation based method (Kernel SHAP). Note that these methods have different costs associate with computing the attribution. Random, A-last and SP-PI are $0$ cost methods as they require minimal/no additional compute. GAE, Inp x Grad and LRP for Transformers require an additional cost (expressed as a multiple of a single inference pass) of $1\times$  as they require some form of backpropagation or gradient computation. 
%Integrated gradients, Kernel SHAP and MP-PI all require us to specify a budget for the number of samples that are used to comptue attributions. 
For Kernel SHAP and MP-PI, we set the number of samples to $2n$, where $n$ is the number of input features (i.e. number of words/sentences). For integrated gradients, we set number of samples to $n$, making the cost $2n$ as each sample requires a forward and backward pass. We use `...' as the mask token for perturbation-based methods.

\subsection{Results}\label{sec:results}
Table~\ref{tab:auc_act} shows the average AUC for each dataset (across 500 examples) from the activation study (higher AUC is better). The best AUC among all the methods is marked in \textbf{bold} and the best AUC among methods with a cost of $\leq 1\times$ is marked with an \underline{underline}. For most datasets, SP-PI provides the best attributions among techniques that have a cost of $\leq 1\times$. Among all techniques, MP-PI provides the best attributions, offering up to a \emph{$10.3\%$ improvement in AUC} compared to the best performing prior work.

Table~\ref{tab:auc_act_inv} shows the average AUC from the inverse activation study (lower AUC is better). Once again, SP-PI provides the best attribution amongst techniques that have a cost of $\leq 1\times$ for most datasets. MP-PI provides significantly better attributions compared to all prior works, offering up to \emph{$57.5\%$ reduction in AUC} over the best performing prior work.
%that are significantly better compared to prior works. %Note that the difference in AUC between MP-PI and prior works is often markedly high (e.g. IMDB, SST-2 and Twitter-Emotion). 
The results from the IAS study highlights the ability of our methods to identify input features that do not support the class being considered for explanation.

We also note that for the same budget ($2n$), MP-PI provides a higher quality attribution compared to Kernel SHAP. This improvement is owed to the higher sample efficiency of MP-PI resulting from the use of intermediate predictions.
%additional samples that are used by MP-PI by interpreting the intermediate predictions as the model's output on masked inputs. 

Due to space limitations, we present the rest of our empirical finings in the Appendix. Appendix~\ref{app:sensitivity} contains the results quantifying the impact of choosing $P'$. Qualitative examples comparing attributions produced by different XAI techniques are provided in Appendix~\ref{app:qualitative_results}. Plots for AS and IAS are provided in Appendix~\ref{app:as_ias_plots}. We compare the similarity between attributions provided by MP-PI with that of Kernel SHAP (with a high sample budget) in Appendix~\ref{app:shap_comparison}. Finally, the limitations of our work are detailed in Appendix~\ref{app:limitations}. Code is provided in the supplementary material.

\section{Conclusion}
We propose a new framework to explain the predictions of decoder-only sequence classification models called \emph{Progressive Inference (PI)}. The key insight of our work is that the intermediate predictions of decoder-only models can be viewed as the predictions of the model on masked versions of the input. We leverage this insight to propose a near zero-cost input attribution technique called Single-Pass PI. We also propose a more sophisticated approach--Multi-Pass PI--that uses multiple inference passes to compute attributions by drawing a connection to SHAP values. Through extensive experiments on a variety of datasets and models we show that SP-PI and MP-PI can significantly outperform prior XAI techniques in terms of the quality of explanations, offering an improvement in AUC of $10.3\%$ for the activation study and $57.5\%$ for the inverse activation study.
%\newpage
\section*{Impact Statement}
Our paper proposes new methods to explain the predictions of decoder-only sequence classification model through input attributions. By providing better attributions, our methods improve the interpretability of ML models, enabling human reviewers to better understand model predictions. Thus, by providing high-quality explanations, our work improves the trustworthiness of ML models, supporting the safe and responsible deployment of AI.
\section*{Acknowledgements}
This paper was prepared for informational purposes by the Artificial Intelligence Research group of JPMorgan Chase \& Co and its affiliates (“J.P. Morgan”) and is not a product of the Research Department of J.P. Morgan.  J.P. Morgan makes no representation and warranty whatsoever and disclaims all liability, for the completeness, accuracy or reliability of the information contained herein.  This document is not intended as investment research or investment advice, or a recommendation, offer or solicitation for the purchase or sale of any security, financial instrument, financial product or service, or to be used in any way for evaluating the merits of participating in any transaction, and shall not constitute a solicitation under any jurisdiction or to any person, if such solicitation under such jurisdiction or to such person would be unlawful. 
\bibliography{references}
\bibliographystyle{icml2024}
\clearpage
\appendix

\section{Proofs}\label{app:proof}

\textbf{\emph{Proposition 1.}} SPPI's attribution $\phi_i = p_i^c - p^c_{i-1}$ satisfies the \emph{local accuracy} property: $p^c_n-p^c_0 = \sum_{i=1}^n \phi_i$.
\begin{proof}[\textbf{Proof.}]
From Eqn.~\ref{eq:3}, we have $\phi_i = p_i^c - p^c_{i-1} $. Expanding $\sum_{i=1}^n \phi_i$, we have
\begin{align}
    \sum_{i=1}^n \phi_i &= \sum_{i=1}^n p_i^c - p^c_{i-1} \\
    \sum_{i=1}^n \phi_i &= (p_1^c - p_0^c) + (p_2^c - p_1^c) + (p_3^c - p_2^c) +..\nonumber\\ 
    &...+ (p_n^c - p_{n-1}^c). 
\end{align}
All the terms except $p_0^c$ and $p_n^c$ cancel out, yielding $\sum_{i=1}^n \phi_i = p_n^c - p_0^c$
\end{proof}

\textbf{\emph{Proposition 2.}} Let $\vec{P}^*=[P^*_1, P^*_2,...,P^*_{n-1}]$ denote the vector representation of the Shapley distribution, where $P^*_i$ denotes the probability of sampling a coalition of size $i$. Let $P^*$ denote the matrix representation of the Shapley distribution consisting of entries $P^*_{ij}$ that indicate the probability of sampling coalitions of size $i$, where $j$ is the last active feature. Then, $P^*_{ij}$ can be expressed in terms of $P^*_i$ as follows: 
\begin{align}\label{eq:prop2}
P^*_{ij} = \begin{cases}
P^*_i \binom{j-1}{i-1}/\binom{n}{i} & \text{if } j \geq i \\
0 & \text{otherwise.}
\end{cases}
\end{align}
\begin{proof}[\textbf{Proof.}]
Since we have $n$ features, the total number of ways in which a subset of $i$ can be formed is $\binom{n}{i}$. If $j\geq i$, the total number of subsets where $j$ is the last active feature is given by $\binom{j-1}{i-1}$. Thus, the probability of selecting a subset of $i$ features, where $j$ is the last active feature is $\binom{j-1}{i-1}/\binom{n}{i}$. Multiplying this with the probability of sampling a coalition of size $i$, we have, $P^*_{ij}=P^*_i\binom{j-1}{i-1}/\binom{n}{i}$ if $j\geq i$. Note that no coalition of size $i$ can be selected such that the index of the last active feature is less than $i$. Thus, $P^*_{ij}=0$ when $j<i$.
\end{proof}

\textbf{\emph{Proposition 3.}} In PI, predictions on an input coalition $S'_{ij}$ (representing the input $\vec{x}'$ in Fig.~\ref{fig:mppi}b), yields additional predictions for coalitions of the form $\{S^\dagger_{kl}\}_{k=1}^i$ i.e. coalitions of sizes $1,2,..,i$ (represented by $D^\dagger_r$ in Fig.~\ref{fig:mppi}e). We can view the distribution of these additional coalitions $S^\dagger_{kl}$ as being conditioned on $S'_{ij}$. Assuming $i,j,k,l\in N$, this conditional distribution is given by
\begin{align}
P^\dagger_{kl|ij} = \begin{cases}
\binom{l-1}{k-1}\binom{j-l}{i-k}/(\binom{j-1}{i-1}i) & \text{if } k \leq i, l\leq j, j\geq i \\
0 & \text{otherwise.}
\end{cases}
\end{align}
\begin{proof}[\textbf{Proof.}]
The total number of coalitions of the form $S'_{ij}$ is given by $\binom{j-1}{i-1}$. For $S'_{ij}$ to yield an intermediate coalition of the form $S^\dagger_{kl}$, we need two conditions to hold: 
\begin{itemize}
\item Feature $l$ to be the $k^{th}$ active feature.
\item There need to be exactly $i-k$ active features after feature $l$.
\end{itemize}
There are $\binom{l-1}{k-1}$ possible ways of satisfying the first condition and $\binom{j-l}{i-k}$ ways of satisfying the second. Thus, totally, there are a total of $\binom{l-1}{k-1}\binom{j-l}{i-k}$ possible coalitions of the form $S'_{ij}$ that satisfy both conditions. Expressed as a fraction of the total number of possible coalitions of the form $S'_{ij}$, this yields $\binom{l-1}{k-1}\binom{j-l}{i-k}/\binom{j-1}{i-1}$. Since we get a total of $i$ intermediate coalitions, we divide by $i$ to obtain the normalized conditional probability  $P^\dagger_{kl|ij} = \binom{l-1}{k-1}\binom{j-l}{i-k}/(\binom{j-1}{i-1}i)$. Note that this probability only holds when $k\leq i, l\leq j$ and $j\geq i$. Under all other conditions, there are no intermediate coalitions that satisfy the conditions above, resulting in $P^\dagger_{kl|ij} =0$.

\end{proof}

\section{Additional Experimental Details}

\subsection{Training Setup}\label{app:training}

We train all models for 10 epochs with a learning rate of $5\times10^{-5}$. We use the Adam optimizer and a batch size of 16. We truncate the inputs when necessary so that it fits within the support input lengths for GPT-2 and Llama-2. For the TrueTeacher dataset, we use the following format in the input: ``[Assertion]: \emph{hypothesis} [Document]: \emph{premise}". Note that putting the \emph{hypothesis} up front allows us to make intermediate predictions on masked versions of the premise.  For LoRA, we use a rank=16, alpha=32 and lora\_dropout=0.1

\subsection{Note on LRP and inp x grad}\label{app:lrp_ixg}
For the LRP and inp x grad methods, we found that taking the l2 norm improves the AUC for both activation and inverse activation studies. This empirical finding is consistent with the results reported in prior work~\cite{atanasova2020diagnostic}. Thus, to have the best performing version of prior work, we use the l2 norm for computing attributions with LPR and inp x grad.

\section{Quantifying MP-PI's Sensitivity to $P'$}\label{app:sensitivity}
A key component of our proposed MP-PI method is finding an optimal $P'$ that results in the distribution of intermediate samples resembling the Shapley distribution. Table~\ref{tab:mppi_shap} quantifies the marginal benefit of choosing this optimal $P'$ over an alternative sampling scheme of directly using the Shapley distribution to sample. For most datasets we find that the optimized sampling scheme provides higher quality explanations, measured in terms of the AUC of the activation and inverse activation studies.

\begin{table}[htb]
\caption{Comparing the performance of MP-PI with the optimal sampling scheme and the default Shapley sampling}
\centering 
\scalebox{0.9}{
\begin{tabular}{l|cc|cc}
\toprule
Dataset &\multicolumn{2}{c|}{AUC Act. ($\uparrow$)}& \multicolumn{2}{c}{AUC Inv. Act. ($\downarrow$)}\\ 
&Opt & Shap & Opt & Shap\\
\midrule
IMDB & \textbf{0.9697}& 0.9696& \textbf{0.6128}& 0.6263\\
SST-2 &\textbf{0.9562}& 0.9461& \textbf{0.4261}&0.4307\\
AG-News &0.8669& \textbf{0.8758}& \textbf{0.5957}&0.5958\\
Twitter-Fin & 0.847& \textbf{0.856}& \textbf{0.67}& 0.6842\\
Twitter-Sentiment & \textbf{0.8866}& 0.8843& \textbf{0.2733}& 0.2779\\
Twitter-Emotion & \textbf{0.932}&0.9292&\textbf{0.2144}& 0.2242\\
TrueTeacher NLI & \textbf{0.921}&0.9198&\textbf{0.3545}& 0.3702\\
\bottomrule
\end{tabular}
}
\label{tab:mppi_shap}
\end{table}

\section{Additional Results}\label{app:additional_results}

\subsection{Qualitative Evaluation of Attributions}\label{app:qualitative_results}

\begin{figure*}[t]
  \centering
  \includegraphics[width=\textwidth]{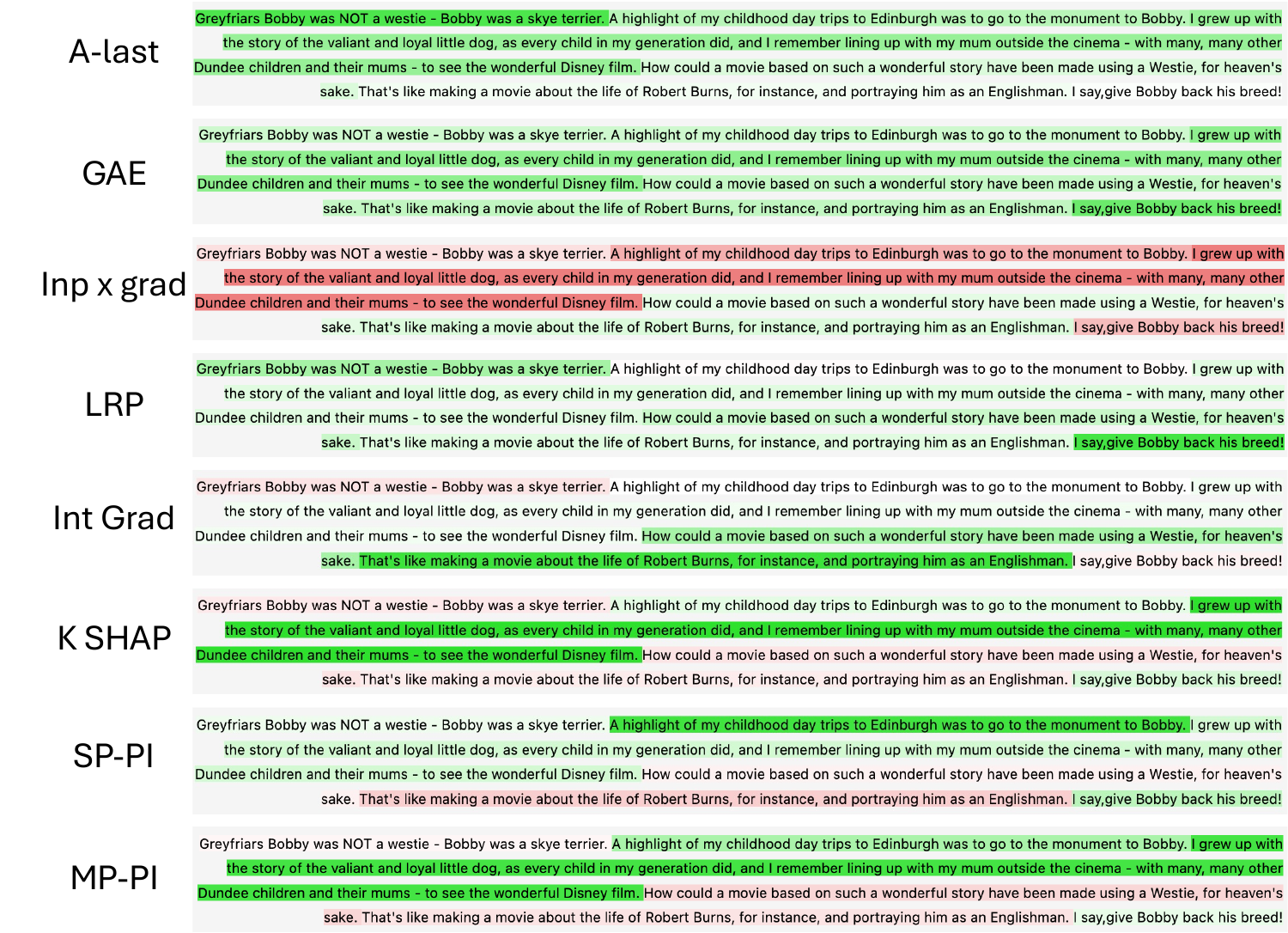}
  \caption{Comparing the attributions provided by different XAI techniques on a mispredicted sample from the IMDB dataset. The above example is a negative movie review (class 0). The model incorrectly classifies this example as a positive review (class 1). Sentences with positive attributions are highlighted in green and sentences with negative attributions in red. Our proposed methods (SP-PI, MP-PI) provide attributions that are consistent with the sentiment of each sentence, while most prior works provide inconsistent attributions. Faithful explanations such as the ones provided by SP-PI and MP-PI allow a human reviewer to easily identify the misprediction.}
  \label{fig:qualitative results1}
\end{figure*}

\begin{figure*}[t]
  \centering
  \includegraphics[width=\textwidth]{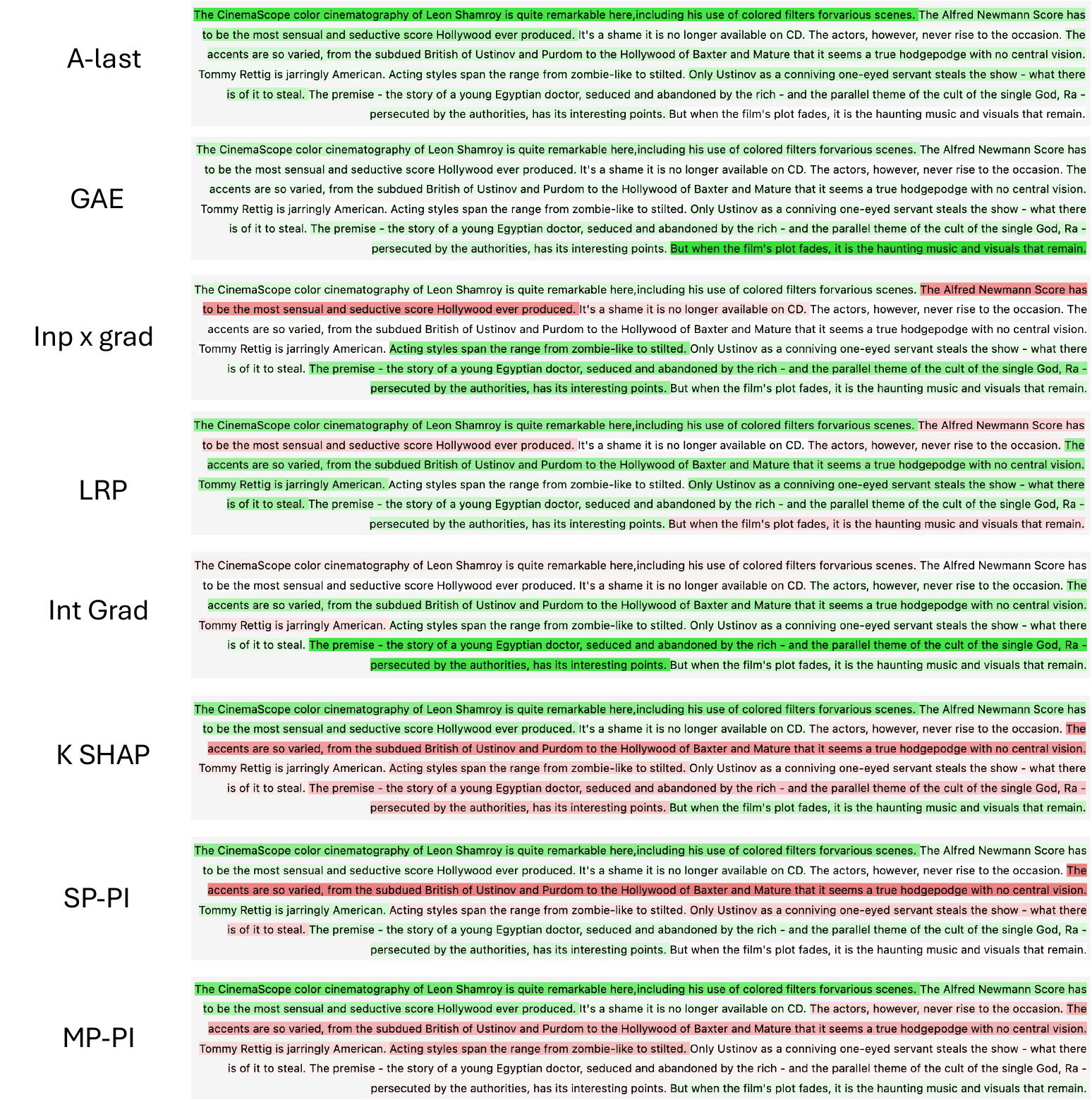}
  \caption{Comparing the attributions provided by different XAI techniques on a mispredicted sample from the IMDB dataset. The above example is a negative movie review (class 0). The model incorrectly classifies this example as a positive review (class 1). Sentences with positive attributions are highlighted in green and sentences with negative attributions in red. Our proposed methods (SP-PI, MP-PI) provide attributions that are consistent with the sentiment of each sentence, while most prior works provide inconsistent attributions. Faithful explanations such as the ones provided by SP-PI and MP-PI allow a human reviewer to easily identify the misprediction.}
  \label{fig:qualitative results2}
\end{figure*}

Fig.~\ref{fig:qualitative results1},~\ref{fig:qualitative results2} compares the attributions produced by different XAI techniques on two mispredicted examples from the IMDB dataset. Note that the example contains a negative movie review, which is mispredicted as a positive review by the model in both cases. We compute the attribution with respect to the predicted class (i.e. the positive class). Attributions are computed at the sentence level. Sentences that support the prediction (i.e. positive sentences) are highlighted in green and ones that don't support the prediction (i.e. negative sentences) are highlighted in red. The shade of red/green indicates the magnitude of the normalized attribution. Note that for a human reviewer to catch this mistake, it is important for the XAI technique to highlight sentences that don't support the prediction. We see that in both cases, A-last and GAE fail to highlight any sentence in red. The attributions provided by inp x grad, LRP and integrated gradients are incorrect as they fail to properly highlight positive and negative sentences\footnote{Just for collecting these qualitative samples, we don't take the l2 norm of the attributions for inp x grad and LRP as taking the norm would result in only positive attributions.}. Only Kernel SHAP, SP-PI and MP-PI provide attributions that are consistent with the sentiment of each sentence. This shows that perturbation based attribution methods such as the ones proposed in this paper provide attributions that are the most helpful in the event of a misprediction. Our quantitative results in Section~\ref{sec:results} support these qualitative findings.

\subsection{Plots for Activation and Inverse Activation Studies}\label{app:as_ias_plots}
\begin{figure*}[t]
  \centering
  \includegraphics[width=0.8\textwidth]{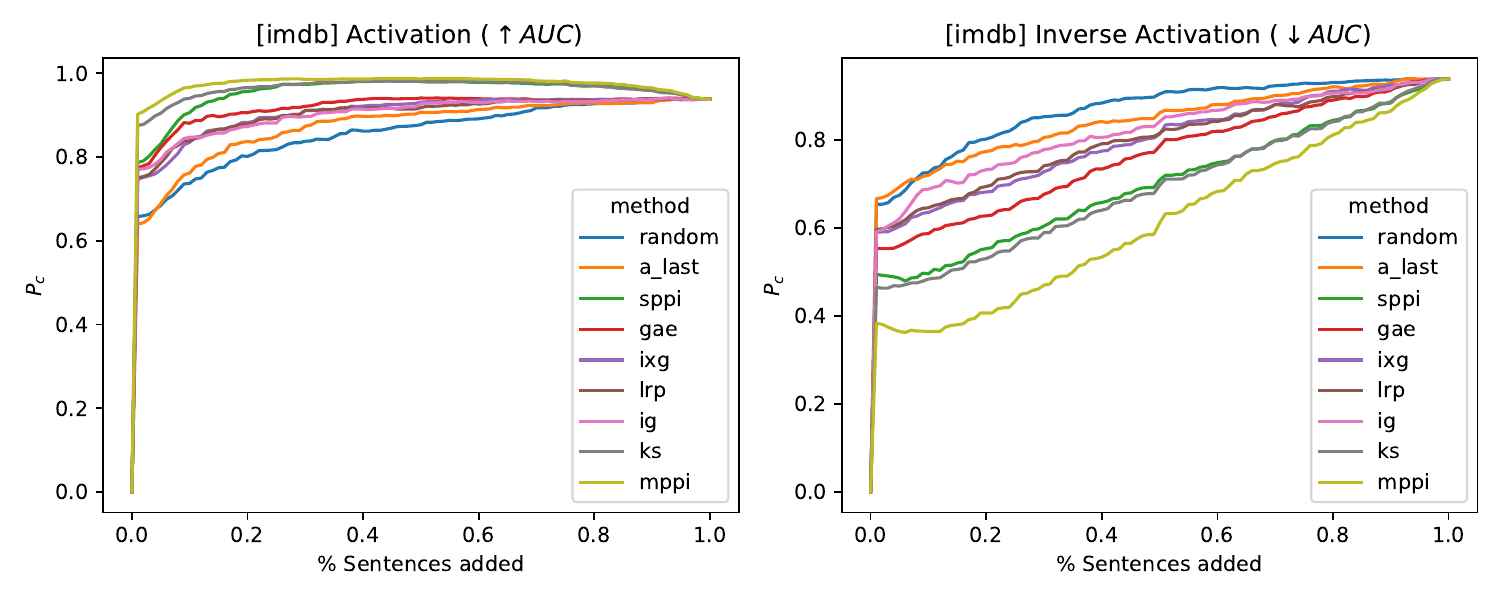}
  \includegraphics[width=0.8\textwidth]{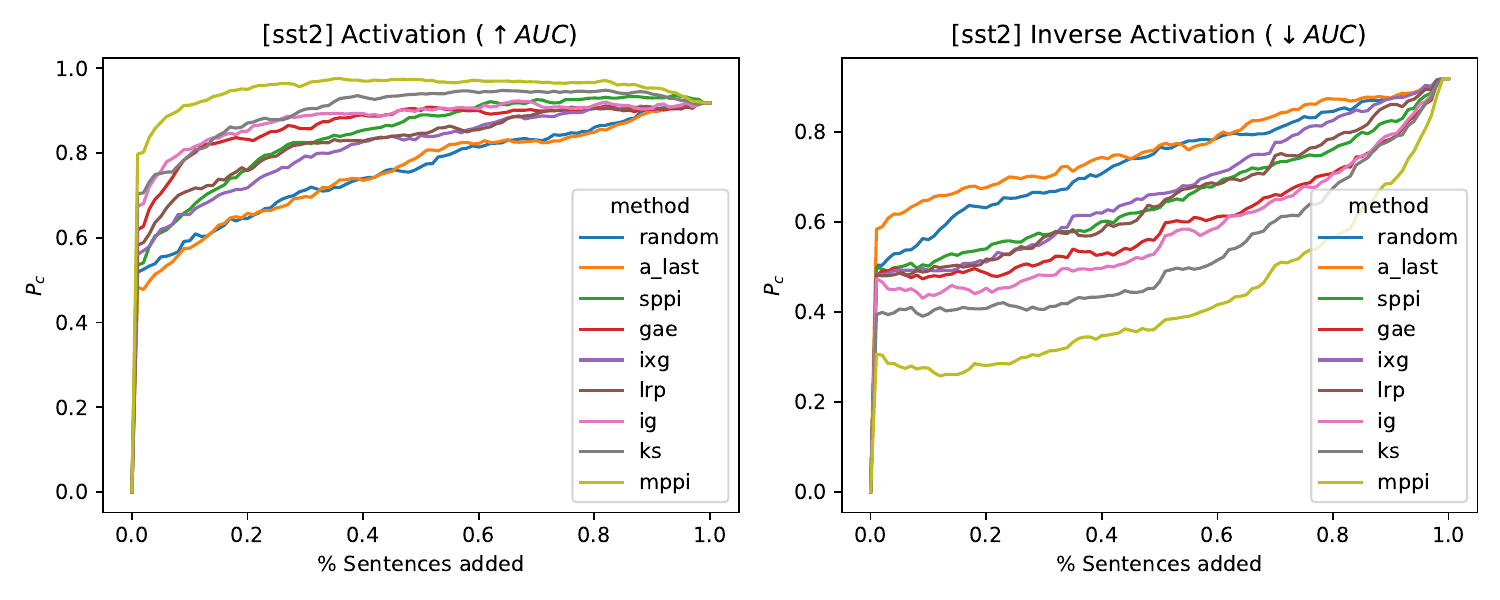}
  \includegraphics[width=0.8\textwidth]{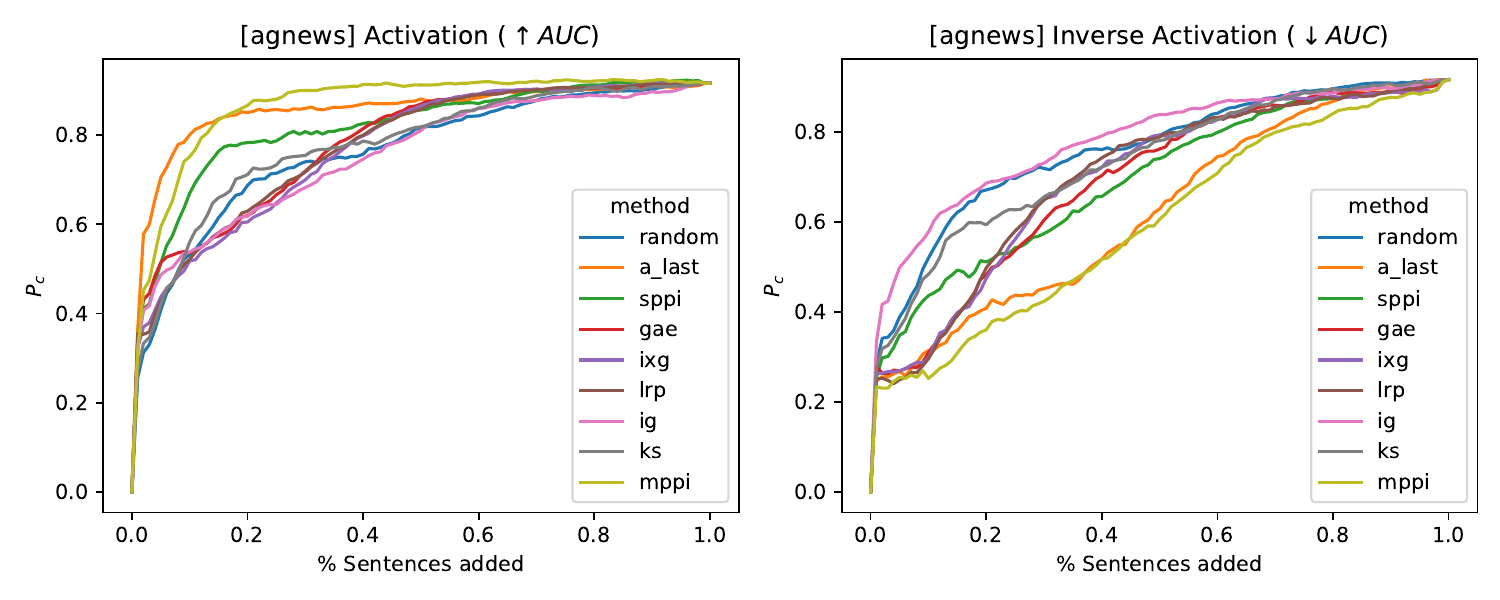}
  \includegraphics[width=0.8\textwidth]{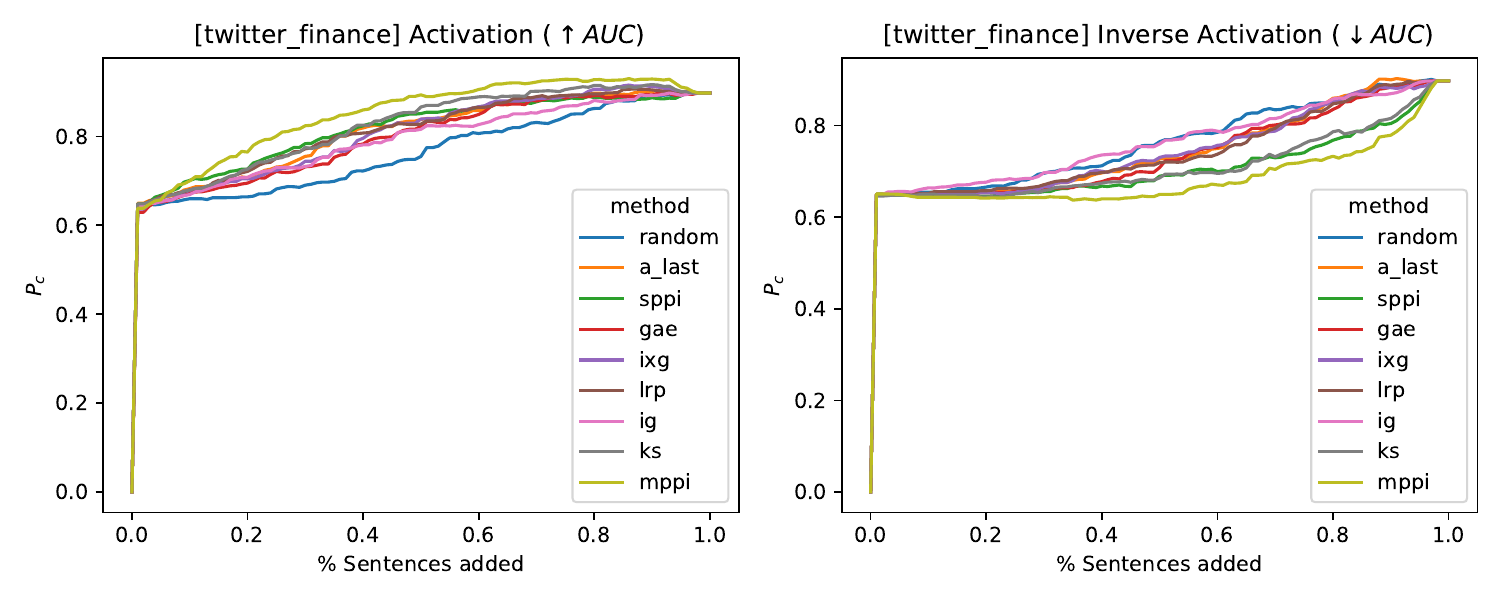}
  \caption{Plots for activation and inverse activation studies.}
  \label{fig:act_inv_act1}
\end{figure*}

\begin{figure*}[t]
  \centering
  \includegraphics[width=0.8\textwidth]{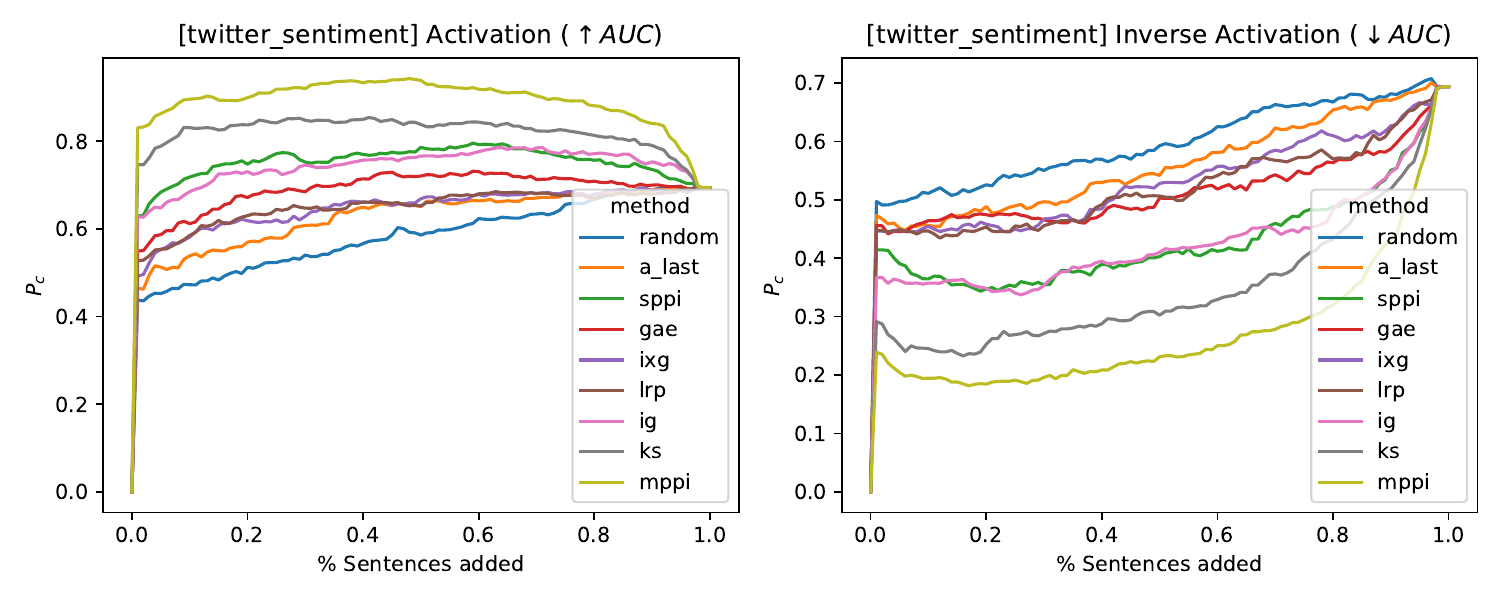}
  \includegraphics[width=0.8\textwidth]{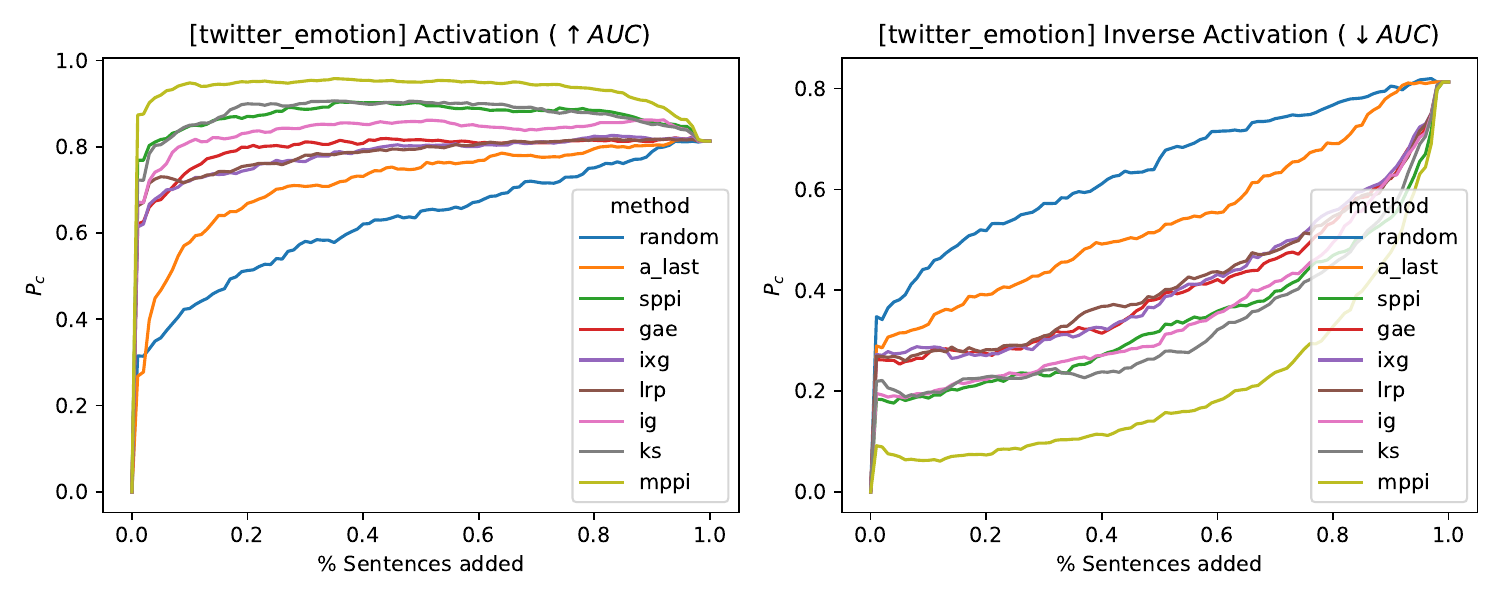}
  \caption{Plots for activation and inverse activation studies.}
  \label{fig:act_inv_act2}
\end{figure*}

Fig.~\ref{fig:act_inv_act1},~\ref{fig:act_inv_act2} show the plots for activation and inverse activation studies.

\subsection{Statistical Significance of Activation and Inverse Activation Studies}\label{app:as_ias_stat}
Table~\ref{tab:auc_act_ci} and Table~\ref{tab:auc_act_inv_ci} list the $95\%$ confidence intervals for the mean AUC reported in Table~\ref{tab:auc_act} and Table~\ref{tab:auc_act_inv} respectively. Note that the width of the confidence intervals is smaller than the magnitude of improvements offered by our proposal over prior works.

\begin{table*}[t]
\caption{95\% Confidence intervals for Activation Study (Table~\ref{tab:auc_act})}
\centering 
\scalebox{0.8}{
\begin{tabular}{l|lllllll}
\toprule
Method       & \rotatebox{\rot}{IMDB} & \rotatebox{\rot}{SST-2} & \rotatebox{\rot}{AG-news} &   \rotatebox{\rot}{Twitter-Fin} & \rotatebox{\rot}{Twitter-Sen} & \rotatebox{\rot}{Twitter-Emo} & \rotatebox{\rot}{TrueTeacher} \\
\midrule
Random & $\pm0.018$ & $\pm0.024$ & $\pm0.022$ & $\pm0.029$ & $\pm0.03$ & $\pm0.028$ & $\pm0.025$ \\
A-Last & $\pm0.017$ & $\pm0.024$ & $\pm0.022$ & $\pm0.026$ & $\pm0.03$ & $\pm0.027$ & $\pm0.025$ \\
SP-PI & $\pm0.008$ & $\pm0.019$ & $\pm0.021$ & $\pm0.026$ & $\pm0.027$ & $\pm0.021$ & $\pm0.019$ \\
GAE & $\pm0.017$ & $\pm0.021$ & $\pm0.022$ & $\pm0.026$ & $\pm0.029$ & $\pm0.027$ & $\pm0.024$ \\
Inp x Grad & $\pm0.018$ & $\pm0.023$ & $\pm0.022$ & $\pm0.025$ & $\pm0.032$ & $\pm0.028$ & $\pm0.025$ \\
LRP for Transformers & $\pm0.018$ & $\pm0.023$ & $\pm0.022$ & $\pm0.025$ & $\pm0.032$ & $\pm0.029$ & $\pm0.025$ \\
Int. Grad & $\pm0.016$ & $\pm0.019$ & $\pm0.023$ & $\pm0.027$ & $\pm0.029$ & $\pm0.024$ & $\pm0.024$ \\
Kernel SHAP & $\pm0.009$ & $\pm0.016$ & $\pm0.021$ & $\pm0.024$ & $\pm0.025$ & $\pm0.021$ & $\pm0.022$ \\
MP-PI & $\pm0.007$ & $\pm0.011$ & $\pm0.019$ & $\pm0.022$ & $\pm0.018$ & $\pm0.015$ & $\pm0.015$ \\
\bottomrule
\end{tabular}
}
\label{tab:auc_act_ci}
\end{table*}

\begin{table*}[t]
\caption{95\% Confidence intervals for Inverse Activation Study (Table~\ref{tab:auc_act_inv})}
\centering 
\scalebox{0.8}{
\begin{tabular}{l|lllllll}
\toprule
Method       & \rotatebox{\rot}{IMDB} & \rotatebox{\rot}{SST-2} & \rotatebox{\rot}{AG-news} &   \rotatebox{\rot}{Twitter-Fin} & \rotatebox{\rot}{Twitter-Sen} & \rotatebox{\rot}{Twitter-Emo} & \rotatebox{\rot}{TrueTeacher} \\
\midrule
Random & $\pm0.017$ & $\pm0.024$ & $\pm0.023$ & $\pm0.029$ & $\pm0.03$ & $\pm0.027$ & $\pm0.026$ \\
A-Last & $\pm0.02$ & $\pm0.024$ & $\pm0.021$ & $\pm0.03$ & $\pm0.029$ & $\pm0.027$ & $\pm0.029$ \\
SP-PI & $\pm0.026$ & $\pm0.029$ & $\pm0.024$ & $\pm0.033$ & $\pm0.031$ & $\pm0.028$ & $\pm0.032$ \\
GAE & $\pm0.018$ & $\pm0.026$ & $\pm0.022$ & $\pm0.031$ & $\pm0.031$ & $\pm0.026$ & $\pm0.033$ \\
Inp x Grad & $\pm0.017$ & $\pm0.024$ & $\pm0.021$ & $\pm0.03$ & $\pm0.029$ & $\pm0.025$ & $\pm0.034$ \\
LRP for Transformers & $\pm0.018$ & $\pm0.024$ & $\pm0.021$ & $\pm0.03$ & $\pm0.03$ & $\pm0.024$ & $\pm0.034$ \\
Int. Grad & $\pm0.02$ & $\pm0.029$ & $\pm0.023$ & $\pm0.03$ & $\pm0.033$ & $\pm0.029$ & $\pm0.032$ \\
Kernel SHAP & $\pm0.026$ & $\pm0.03$ & $\pm0.023$ & $\pm0.033$ & $\pm0.03$ & $\pm0.028$ & $\pm0.031$ \\
MP-PI & $\pm0.026$ & $\pm0.029$ & $\pm0.024$ & $\pm0.035$ & $\pm0.029$ & $\pm0.021$ & $\pm0.031$ \\
\bottomrule
\end{tabular}
}
\label{tab:auc_act_inv_ci}
\end{table*}

\subsection{Similarity with SHAP values}\label{app:shap_comparison}
\begin{figure}[ht]
  \centering
  \includegraphics[width=1.0\columnwidth]{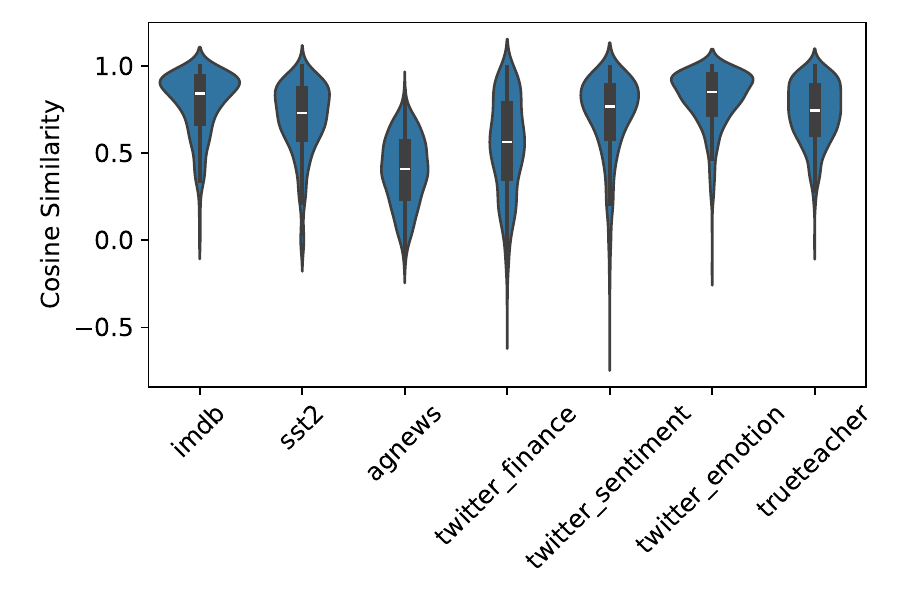}
  \caption{Distribution of cosine similarities between Kernel SHAP and MP-PI attributions. For most datasets, we see a high cosine similarity, indicating that the attributions produced by MP-PI indeed approximates SHAP values.}
  \label{fig:shap_agreement}
\end{figure}

Our work uses intermediate predictions to compute input attributions that approximate SHAP values. To validate this claim, we compare the attributions produced by our method with those obtained by running Kernel SHAP with a very high sample budget (budget =$16n$, where $n$ represents the number of features). We plot the distribution of cosine similarity between the attributions to understand how closely the two attributions match up. The results are shown in Fig.~\ref{fig:shap_agreement}. We find that there is a high degree of similarity between the attributions provided by MP-PI and that of Kernel SHAP for most datasets.

\section{Limitations}\label{app:limitations}
While our methods are capable of providing high quality attributions, there are some limitations that need to be considered when using them in practice.
\begin{itemize}[noitemsep, leftmargin=*, topsep=0pt]
    \item \emph{Validity of masked inputs:} Our methods compute attributions by considering the prediction of the model on masked/perturbed versions of the input. This assumes that masked versions of the input are valid inputs to the model.
    \item \emph{Difference with SHAP values:} The attributions computed by MP-PI differ from SHAP values due to two key reasons. First, the samples obtained in PI are correlated due to the masked attention mechanism and second, the intermediate predictions may not accurately reflect the prediction of the model on the equivalent masked input. 
\end{itemize}

\end{document}